\documentclass{bmvc2k}
\usepackage{etoolbox}
\usepackage{times}
\usepackage{graphicx}
\usepackage{color}
\usepackage{amsmath}
\usepackage{amssymb}
\usepackage{multirow}
\usepackage{tabularx,booktabs}
\definecolor{captioncolor}{rgb}{0,0,.4}
\usepackage[font={color=captioncolor},skip=0pt]{caption}
\usepackage{pifont}
\usepackage{xurl}
\usepackage[percent]{overpic}
\usepackage{soul}
\usepackage[subrefformat=parens]{subcaption}

\newtoggle{bmvcfinal}
\newtoggle{bmvcarxiv}
\togglefalse{bmvcfinal}
\togglefalse{bmvcarxiv}
\toggletrue{bmvcfinal}
\toggletrue{bmvcarxiv}

\newcommand\blfootnote[1]{%
	\begingroup
	\renewcommand\thefootnote{}\footnote{#1}%
	\addtocounter{footnote}{-1}%
	\endgroup
}

\title{USB: Universal-Scale Object Detection Benchmark}

\addauthor{Yosuke Shinya$^*$}{https://shinya7y.github.io/}{1}

\addinstitution{
 independent researcher\\
 Tokyo, Japan
}

\runninghead{Yosuke Shinya}{USB: Universal-Scale Object Detection Benchmark}

\def\eg{\emph{e.g}\bmvaOneDot}

\def\etc{\emph{etc}\bmvaOneDot}
\def\vs{\emph{vs}\bmvaOneDot}

\newcommand{\Univs}{UniverseNets\xspace}
\newcommand{\OurOrig}{UniverseNet\xspace}

\newcommand{\OurGFL}{UniverseNet$+$GFL\xspace}

\newcommand{\OurAugustD}{UniverseNet-20.08d\xspace}
\newcommand{\OurAugust}{UniverseNet-20.08\xspace}

\newcommand{\OurAugustS}{UniverseNet-20.08s\xspace}

\iftoggle{bmvcfinal}{\newcommand{\UnivRepo}{\url{https://github.com/shinya7y/UniverseNet}\xspace}}
{\newcommand{\UnivRepo}{our GitHub repository and in the Supplementary Material\xspace}}
\iftoggle{bmvcfinal}{\newcommand{\WaymoCOCO}{\url{https://github.com/shinya7y/WaymoCOCO}\xspace}}
{\newcommand{\WaymoCOCO}{Our GitHub repository (anonymized for review).\xspace}}
\iftoggle{bmvcfinal}{\newcommand{\mangaapi}{\url{https://github.com/shinya7y/manga109api}\xspace}}
{\newcommand{\mangaapi}{Our GitHub repository (anonymized for review).\xspace}}
\iftoggle{bmvcfinal}{}
{}

\newcommand{\ATSEPC}{ATSEPC\xspace}

\def\AppendixSection{Supp.\xspace}

\newcommand{\cm}{\ding{51}}%
\newcommand{\xm}{\ding{55}}%

\newcommand{\APS}{AP$_\textit{S}$\xspace}
\newcommand{\APM}{AP$_\textit{M}$\xspace}
\newcommand{\APL}{AP$_\textit{L}$\xspace}

\newcommand{\TB}{\textbf}
\newcommand{\K}{\,k\xspace}
\newcommand{\M}{\,M\xspace}
\newcommand{\Hz}{\,Hz\xspace}

\newcommand{\Pascal}{\textsc{Pascal}\xspace}
\newcommand{\Mangas}{Manga109-s\xspace}
\newcommand{\MangasAbbr}{M109s\xspace}

\AtBeginDocument{
	\abovedisplayskip     =0.3\abovedisplayskip
	\abovedisplayshortskip=0.3\abovedisplayshortskip
	\belowdisplayskip     =0.3\belowdisplayskip
	\belowdisplayshortskip=0.3\belowdisplayshortskip
}

\begin{document}

\maketitle

\iftoggle{bmvcfinal}{
\blfootnote{
	\hspace{-19pt} $*$This work was done independently of the author's employer.}
\vspace{-15pt}
}

\begin{center}
	\begin{minipage}[b]{0.32\hsize}
		\centering
		\includegraphics[width=\textwidth,height=0.707376\textwidth]{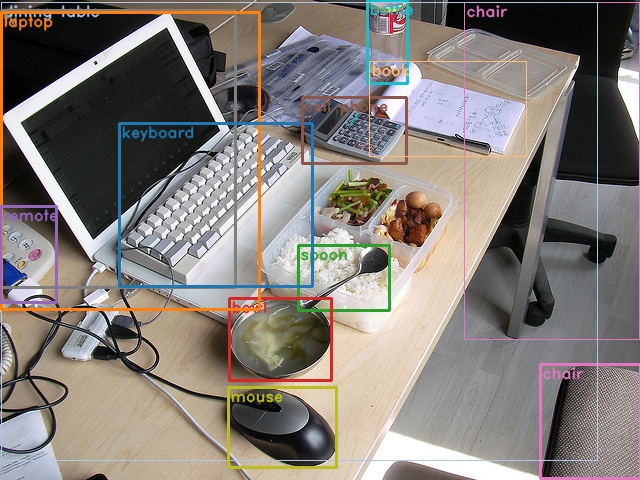}
		\label{fig:teaser_a}
	\end{minipage}
	\hfill
	\begin{minipage}[b]{0.339540508\hsize}
		\centering
		\includegraphics[width=\textwidth]{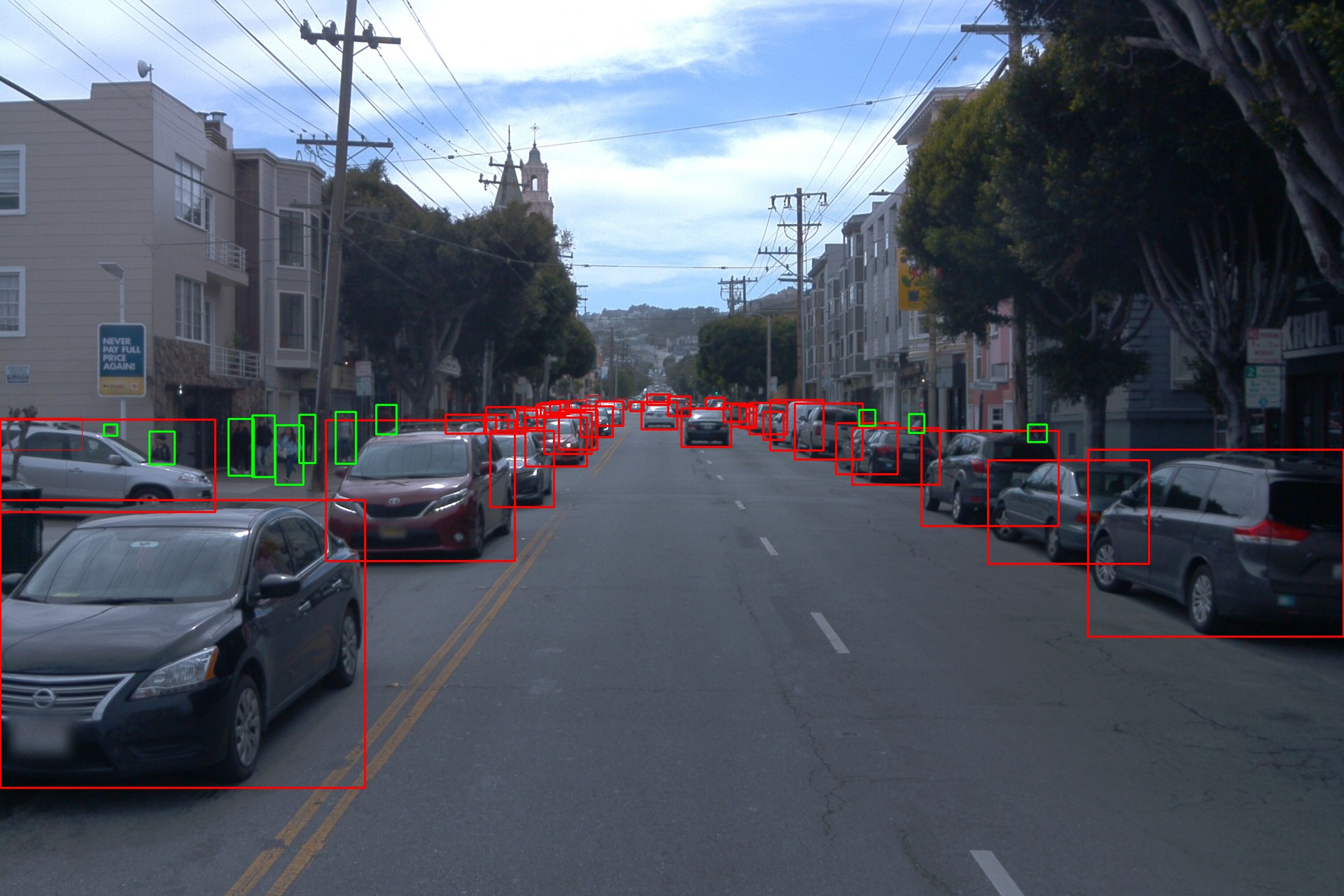}
		\label{fig:teaser_b}
	\end{minipage}
	\hfill
	\begin{minipage}[b]{0.32\hsize}
		\centering
		\begin{overpic}[width=\textwidth]{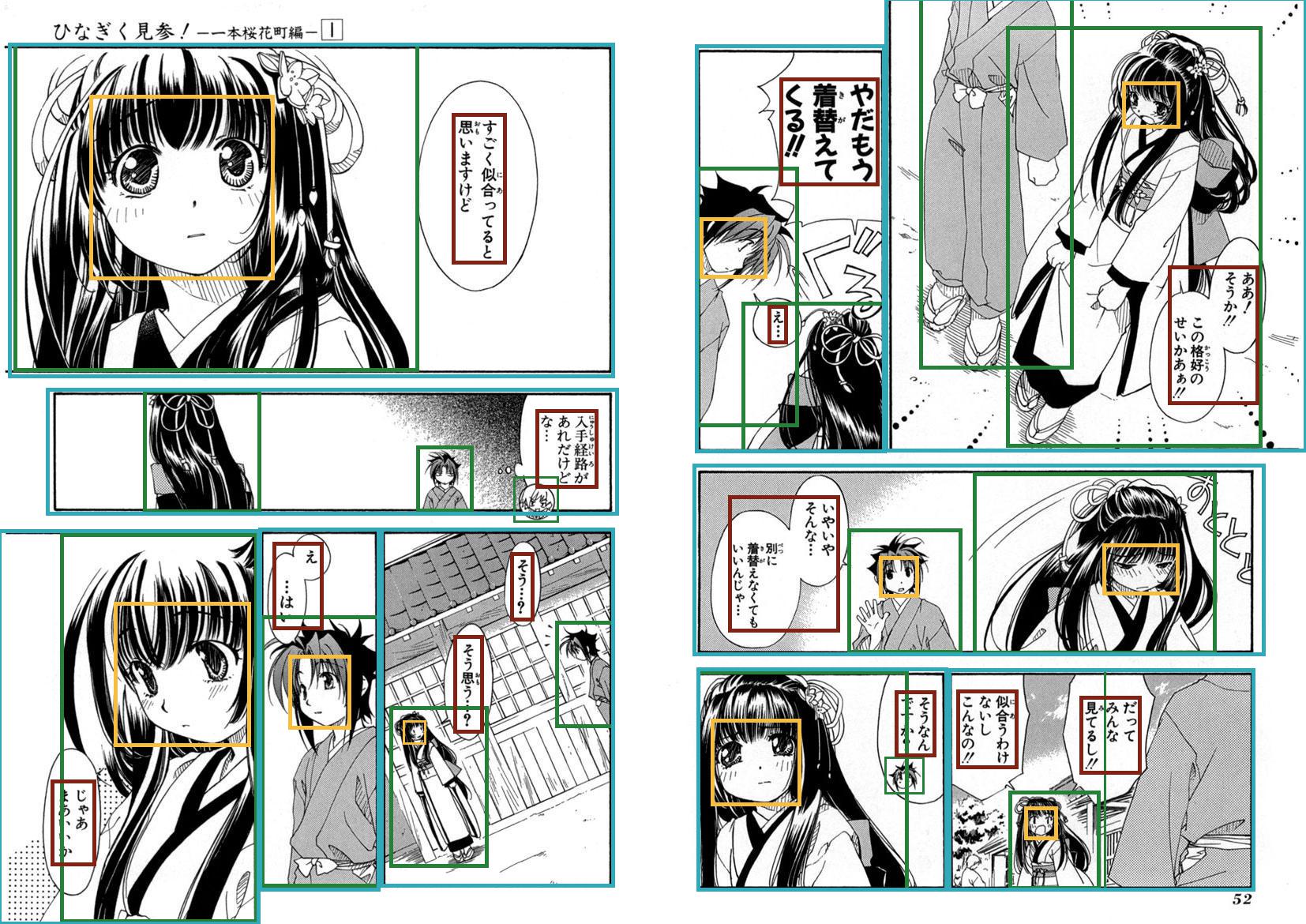}
			\put( 34, -1.5){\scalebox{0.7}{\scriptsize{Hinagiku Kenzan\textit{!} \copyright Minene Sakurano}}}
		\end{overpic}
		\label{fig:teaser_c}
	\end{minipage}
	\vspace{-3mm}
	\captionof{figure}{
		Universal-scale object detection.
		For realizing human-level perception,
		object detection systems must detect both tiny and large objects,
		even if they are out of natural image domains.
		To this end,
		we introduce the \textit{Universal-Scale object detection Benchmark (USB)} that
		consists of the COCO dataset (left), Waymo Open Dataset (middle), and \Mangas dataset (right).
	}
	\label{fig:teaser}
\end{center}

\begin{abstract}
Benchmarks, such as COCO, play a crucial role in object detection.
However, existing benchmarks are insufficient in scale variation, and their protocols are inadequate for fair comparison.
In this paper, we introduce the Universal-Scale object detection Benchmark (USB).
USB has variations in object scales and image domains
by incorporating COCO with the recently proposed Waymo Open Dataset and Manga109-s dataset.
To enable fair comparison and inclusive research, we propose training and evaluation protocols.
They have multiple divisions for training epochs and evaluation image resolutions, like weight classes in sports,
and compatibility across training protocols, like the backward compatibility of the Universal Serial Bus.
Specifically, we request participants to report results with not only higher protocols (longer training) but also lower protocols (shorter training).
Using the proposed benchmark and protocols, we conducted extensive experiments using 15 methods and found weaknesses of existing COCO-biased methods.
The code is available at \UnivRepo.
\end{abstract}

\section{Introduction}
\label{sec:introduction}

Humans can detect various objects. See Figure~\ref{fig:teaser}.
One can detect
close equipment in everyday scenes,
far vehicles in traffic scenes,
and texts and persons in manga (Japanese comics).
If computers can automatically detect various objects,
they will yield significant benefits to humans.
For example, they will
help impaired people and the elderly,
save lives by autonomous driving,
and provide safe entertainment during pandemics by automatic translation.

Researchers have pushed the limits of object detection systems
by establishing datasets and benchmarks~\cite{object_detection_survey_Liu_IJCV2020}.
One of the most important milestones is \Pascal VOC~\cite{PASCALVOC_IJCV2015}.
It has enabled considerable research on object detection,
leading to the success of deep learning-based methods and successor datasets such as ImageNet~\cite{ImageNet_IJCV2015} and COCO~\cite{COCO_ECCV2014}.
Currently, COCO serves as \textit{the} standard dataset and benchmark for object detection
because it has several advantages over \Pascal VOC~\cite{PASCALVOC_IJCV2015}.
COCO contains more images, categories, and objects (especially small objects) in their natural context~\cite{COCO_ECCV2014}.
Using COCO, researchers can develop and evaluate methods for multi-scale object detection.
However, the current object detection benchmarks, especially COCO, have the following two problems.

\textbf{Problem 1: Variations in object scales and image domains remain limited.}
To realize human-level perception,
computers must handle various object scales and image domains as humans can.
Among various domains~\cite{UniversalObjectDetection_CVPR2019},
the traffic and artificial domains have extensive scale variations (see Sec.~\ref{sec:usb_datasets}).
COCO is far from covering them.
Nevertheless, the current computer vision community is overconfident in COCO results.
For example, most studies on state-of-the-art methods in 2020
only report COCO results~\cite{ATSS_CVPR2020, SEPC_CVPR2020, PAA_ECCV2020, GFL_NeurIPS2020, RepPointsv2_NeurIPS2020, RelationNet2_NeurIPS2020} or
those for bounding box object detection~\cite{EfficientDet_CVPR2020, SpineNet_CVPR2020, YOLOv4_2020, DetectoRS_2020}.
Readers cannot assess whether these methods are specialized for COCO or generalizable to other datasets and domains.

\textbf{Problem 2: Protocols for training and evaluation are not well established.}
There are standard experimental settings for the COCO benchmark~\cite{Detectron2018, MMDetection, FPN_CVPR2017, RetinaNet_ICCV2017, FCOS_ICCV2019, ATSS_CVPR2020, GFL_NeurIPS2020}.
Many studies train detectors within 24 epochs
using a learning rate of 0.01 or 0.02
and evaluate them on images within 1333$\times$800.
These settings are not obligations but non-binding agreements for fair comparison.
Some studies do not follow the settings for accurate and fast detectors\footnote{YOLOv4 was trained for 273 epochs~\cite{YOLOv4_2020},
DETR for 500 epochs~\cite{DETR_ECCV2020},
EfficientDet-D6 for 300 epochs~\cite{EfficientDet_CVPR2020},
and EfficientDet-D7x for 600 epochs~\cite{EfficientDet_arXiv}.
SpineNet uses a learning rate of 0.28~\cite{SpineNet_CVPR2020},
and YOLOv4 uses a searched learning rate of 0.00261~\cite{YOLOv4_2020}.
EfficientDet finely changes the image resolution from 512$\times$512 to 1536$\times$1536~\cite{EfficientDet_CVPR2020}.}.
Their abnormal and scattered settings hinder the assessment of the most suitable method.
Furthermore,
by ``buying stronger results''~\cite{GreenAI_CACM2020},
they build a barrier for those without considerable funds to develop and train detectors.

This study makes the following two contributions to resolve the problems.

\textbf{Contribution 1:}
We introduce the \textit{Universal-Scale object detection Benchmark (USB)} that consists of three datasets.
In addition to COCO, we selected the Waymo Open Dataset~\cite{WaymoOpenDataset_CVPR2020} and \Mangas~\cite{Manga109_Matsui_MTAP2017, Manga109_Aizawa_IEEEMM2020} to cover various object scales and image domains.
They are the largest public datasets in their domains and enable reliable comparisons.
To the best of our knowledge,
USB is the first benchmark beyond COCO that evaluates finer scale-wise metrics across multiple domains.
We conducted extensive experiments using 15 methods and found weaknesses of existing COCO-biased methods.

\textbf{Contribution 2:}
We established the \textit{USB protocols} for fair training and evaluation,
inspired by weight classes in sports and the backward compatibility of the Universal Serial Bus.
Specifically, USB protocols enable fair and easy comparisons
by defining multiple divisions for training epochs and evaluation image resolutions.
Furthermore, we introduce compatibility across training protocols
by requesting participants to report results with not only higher protocols (longer training) but also lower protocols (shorter training).
To the best of our knowledge,
our training protocols are the first ones that allow for both fair comparisons with shorter training and strong results with longer training.
Our protocols promote inclusive, healthy, and sustainable object detection research.

\section{Related Work}
\label{sec:related_work}

\noindent
\textbf{Multi-scale object detection.}
Detecting multi-scale objects is a fundamental challenge in object detection~\cite{object_detection_survey_Liu_IJCV2020, UniversalObjectDetection_ZhaoweiCai_2019}.
Various components have been improved, including
backbones and modules~\cite{Inception_CVPR2015, ResNet_CVPR2016, BagOfTricks_Classification_CVPR2019, Res2Net_TPAMI2020, DCN_ICCV2017},
necks~\cite{FPN_CVPR2017, SEPC_CVPR2020, EfficientDet_CVPR2020, DyHead_CVPR2021},
heads and training sample selection~\cite{Faster_R-CNN_NIPS2015, SSD_ECCV2016, ATSS_CVPR2020}, and
multi-scale training and testing~\cite{Rowley_PAMI1998, SNIP_Singh_CVPR2018, ATSS_CVPR2020}
(see \AppendixSection~\ref{sec:details_related_work} for details).
Unlike most prior studies,
we analyzed their methods across various object scales and image domains through the proposed benchmark.

\noindent
\textbf{Single-domain benchmarks.}
There are numerous object detection benchmarks that specialize in a specific domain or consider natural images as a single generic domain.
For specific (category) object detection,
recent benchmarks such as WIDER FACE~\cite{WIDERFACE_CVPR2016} and TinyPerson~\cite{TinyPerson_WACV2020} contain tiny objects.
For autonomous driving,
KITTI~\cite{KITTI_CVPR2012} and Waymo Open Dataset~\cite{WaymoOpenDataset_CVPR2020}
mainly evaluate three categories (car, pedestrian, and cyclist) in their leaderboards.
For generic object detection, \Pascal VOC~\cite{PASCALVOC_IJCV2015} and COCO~\cite{COCO_ECCV2014} include 20 and 80 categories, respectively.
The number of categories has been further expanded by recent benchmarks, such as Open Images~\cite{OpenImagesDataset_IJCV2020}, Objects365~\cite{Objects365_ICCV2019}, and LVIS~\cite{LVIS_CVPR2019}.
The above datasets comprise photographs,
whereas Clipart1k, Watercolor2k, Comic2k~\cite{CrossDomainDetection_Inoue_CVPR2018}, and \Mangas~\cite{Manga109_Matsui_MTAP2017, Manga109_Aizawa_IEEEMM2020}
comprise artificial images.
Although Waymo Open Dataset~\cite{WaymoOpenDataset_CVPR2020} and \Mangas~\cite{Manga109_Matsui_MTAP2017, Manga109_Aizawa_IEEEMM2020} have extensive scale variations (see Sec.~\ref{sec:usb_datasets}), 
scale-wise metrics have not been evaluated~\cite{WaymoOpenDataset_CVPR2020, Manga109_detection_Ogawa_2018}.
Unlike the above benchmarks, our USB consists of multiple domains and contains many instances in both photographs and artificial images,
and we can evaluate the generalization ability of methods.

\noindent
\textbf{Cross-domain benchmarks.}
To avoid performance drops in target domains without labor-intensive annotations,
many studies have tackled domain adaptation of object detection~\cite{DomainAdaptationDetectors_arXiv2021}.
Some datasets have been proposed for this setting~\cite{Sim10k_ICRA2017, CrossDomainDetection_Inoue_CVPR2018}.
Typically, there is a strong constraint to share a label space.
Otherwise, special techniques are needed for training, architectures, unified label spaces~\cite{UniDet_Zhao_ECCV2020, UniDet_Zhou_CVPR2022}, and partial or open-set domain adaptation~\cite{DomainAdaptationDetectors_arXiv2021}.
In contrast, we focus on fully supervised object detection, which allows us to analyze many standard detectors.

\noindent
\textbf{Multi/universal-domain benchmarks.}
Even if target datasets have annotations for training,
detectors trained and evaluated on a specific dataset may perform worse on other datasets or domains.
To address this issue, some benchmarks consist of multiple datasets.
In the Robust Vision Challenge (RVC) 2020~\cite{RVC2020}, detectors were evaluated on three datasets in the natural and traffic image domains.
A few studies have explored the two domains by enriching RVC~\cite{UniDet_Zhou_CVPR2022} or making a unique combination~\cite{UniDet_Zhao_ECCV2020}, although they focus on methods for unified detectors.
For \textit{universal-domain} object detection, the Universal Object Detection Benchmark (UODB)~\cite{UniversalObjectDetection_CVPR2019} comprises 11 datasets in the natural, traffic, aerial, medical, and artificial image domains.
Although it is suitable for evaluating detectors in various domains, variations in object scales are limited.
Unlike UODB, our USB focuses on \textit{universal-scale} object detection.
The datasets in USB contain more instances, including tiny objects, than the datasets used in UODB.

\noindent
\textbf{Criticism of experimental settings.}
For fair, inclusive, and efficient research,
many studies have criticized experimental settings (\eg, \cite{GreenAI_CACM2020, MetricLearningRealityCheck_ECCV2020}).
These previous studies do not propose fair and practical protocols for object detection benchmarks.
As discussed in Sec.~\ref{sec:introduction},
the current object detection benchmarks allow extremely unfair settings (\eg, 25$\times$ epochs).
We resolved this problem by establishing protocols for fair training and evaluation.

\section{Benchmark Protocols of USB}
\label{sec:usb}

Here, we present the principle, datasets, protocols, and metrics of USB.
See \AppendixSection~\ref{sec:details_protocols} for additional information.

\subsection{Principle}
\label{sec:usb_principle}

We focus on the \textit{Universal-Scale Object Detection (USOD)} task
that aims to detect various objects in terms of object scales and image domains.
Unlike separate discussions for multi-scale object detection (Sec.~\ref{sec:related_work}) and universal (-domain) object detection~\cite{UniversalObjectDetection_CVPR2019},
USOD does not ignore the relation between scales and domains (Sec.~\ref{sec:usb_datasets}).

For various applications and users, benchmark protocols should cover 
from short to long training and from small to large test scales.
On the other hand, they should not be scattered for meaningful benchmarks.
To satisfy the conflicting requirements, we define multiple divisions for training epochs and evaluation image resolutions.
Furthermore, we urge participants who have access to extensive computational resources
to report results with standard training settings.
This request enables fair comparison and allows many people to develop and compare object detectors.

\subsection{Definitions of Object Scales}
\label{sec:object_scale_definitions}

Following
\cite{TinyPerson_WACV2020}, we consider two types of object scales.
The absolute scale is calculated as $\sqrt{wh}$,
where $w$ and $h$ denote the object's width and height, respectively.
The relative scale is calculated as $\sqrt{\frac{wh}{WH}}$,
where $W$ and $H$ denote the image's width and height, respectively.

\subsection{Datasets}
\label{sec:usb_datasets}

\begin{table}
\begin{minipage}[c]{0.30\hsize}
	\centering
	\includegraphics[width=1.055\linewidth]{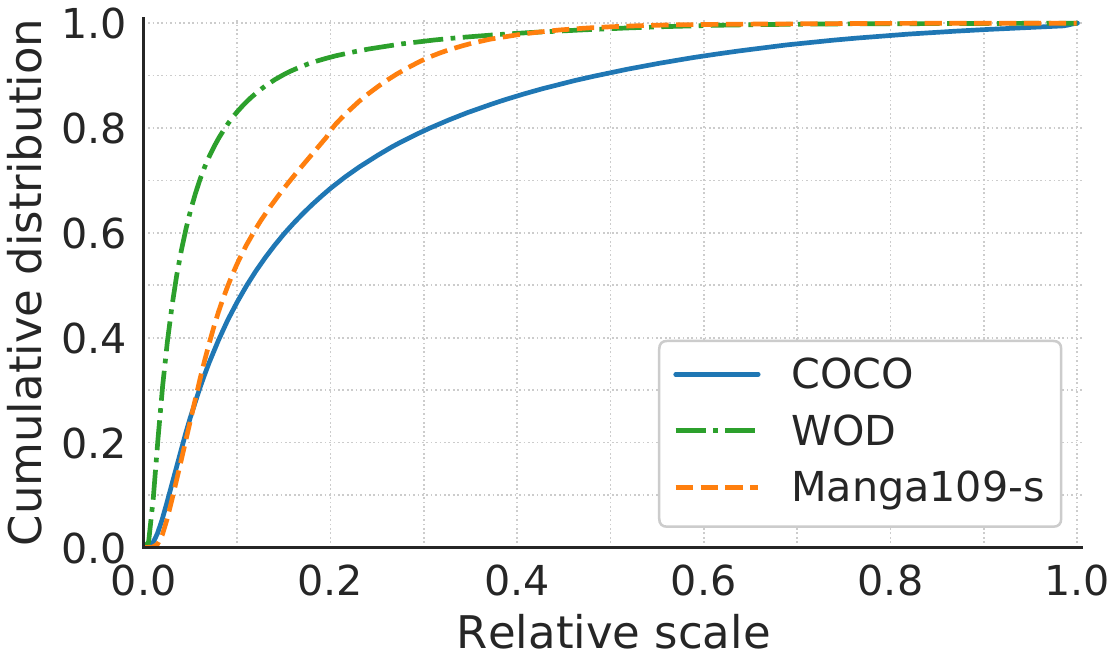}
	\captionof{figure}{
		Distributions of objects' relative scale~\cite{SNIP_Singh_CVPR2018, Waymo2d_1st_2020}.
		USB covers diverse scale variations.
	}
	\label{fig:instance_scale_distribution}
\end{minipage}\hfill
\begin{minipage}[c]{0.67\hsize}
	\setlength{\tabcolsep}{0.9mm}
	\renewcommand\arraystretch{0.85}
	\begin{center}
		\scalebox{0.75}{\begin{tabular}{llcccc}
			\toprule
			Benchmark                                                      & Dataset                                                             &                 Boxes                 &   Images   &    B/I    &      Scale variation$^\dagger$       \\ \midrule
			\multirow{3}{*}{USB (Ours)}                                    & COCO~\cite{COCO_ECCV2014}                                           & \TB{897\K} \small{(\TB{3.1}$\times$)} & \TB{123\K} & \TB{7.3}  &      88.8 \small{(1.0$\times$)}      \\
			                                                               & WOD~\cite{WaymoOpenDataset_CVPR2020} v1.2 \texttt{f0}               & \TB{1.0\M} \small{(\TB{29}$\times$)}  & \TB{100\K} & \TB{10.0} & \TB{96.7} \small{(\TB{5.8}$\times$)} \\
			                                                               & \Mangas~\cite{Manga109_Matsui_MTAP2017, Manga109_Aizawa_IEEEMM2020} & \TB{401\K} \small{(\TB{63}$\times$)}  & \TB{8.2\K} & \TB{49.2} & \TB{28.6} \small{(\TB{1.5}$\times$)} \\ \midrule
			\multirow{3}{*}{UODB~\cite{UniversalObjectDetection_CVPR2019}} & COCO~\cite{COCO_ECCV2014} \texttt{val2014}                          &                 292\K                 &    41\K    &    7.2    &              \TB{89.6}               \\
			                                                               & KITTI~\cite{KITTI_CVPR2012}                                         &                 35\K                  &   7.5\K    &    4.7    &                 16.6                 \\
			                                                               & Comic2k~\cite{CrossDomainDetection_Inoue_CVPR2018}                  &                 6.4\K                 &   2.0\K    &    3.2    &                 19.1                 \\ \bottomrule
		\end{tabular}}
	\end{center}
	\vspace{-3mm}
	\caption{
		Statistics of datasets in USB
		and counterpart datasets in UODB~\cite{UniversalObjectDetection_CVPR2019}.
		Values are based on publicly available annotations.
		B/I: Average number of boxes per image.
		$\dagger$: Calculated by the ratio of the 99 percentile to 1 percentile of relative scale.
	}
	\label{table:dataset_stats}
\end{minipage}
\end{table}

To establish USB, we selected the COCO~\cite{COCO_ECCV2014}, Waymo Open Dataset (WOD)~\cite{WaymoOpenDataset_CVPR2020}, and \Mangas (\MangasAbbr)~\cite{Manga109_Matsui_MTAP2017, Manga109_Aizawa_IEEEMM2020}.
WOD and \MangasAbbr are the largest public datasets with many small objects in the traffic and artificial domains, respectively.
Object scales in these domains vary significantly with distance and viewpoints, unlike those in the medical and aerial domains\footnote{Aerial datasets contain abundant small objects but scarce large ones (see Table 4 in~\cite{DOTA2_TPAMI2021}).
WOD has larger scale variation by distance variation,
where 1\% of objects are larger than $1/4$ of the image area.}.
USB covers diverse scale variations qualitatively (Figure~\ref{fig:teaser}) and quantitatively (Figure~\ref{fig:instance_scale_distribution}).
As shown in Table~\ref{table:dataset_stats},
these datasets contain more instances and larger scale variations~\cite{SNIP_Singh_CVPR2018}
than their counterpart datasets in UODB~\cite{UniversalObjectDetection_CVPR2019}.
USOD needs to evaluate detectors on datasets with many instances
because more instances enable more reliable comparisons of scale-wise metrics.

For the first dataset, we adopted the COCO dataset~\cite{COCO_ECCV2014}.
COCO contains natural images of everyday scenes collected from the Internet.
Annotations for 80 categories are used in the benchmark.
As shown in Figure~\ref{fig:teaser} (left),
object scales mainly depend on categories and distance.
Although COCO contains objects smaller than those of \Pascal VOC~\cite{PASCALVOC_IJCV2015},
objects in everyday scenes (especially indoor scenes) are relatively large.
Since COCO is the current standard dataset for multi-scale object detection,
we adopted the same training split \texttt{train2017}
as the COCO benchmark
to eliminate the need for retraining across benchmarks.
We adopted the \texttt{val2017} split (also known as \texttt{minival}) as the test set.

For the second dataset, we adopted the WOD,
which is a large-scale, diverse dataset for autonomous driving~\cite{WaymoOpenDataset_CVPR2020}
with many annotations for tiny objects (Figure~\ref{fig:instance_scale_distribution}).
The images were recorded using five high-resolution cameras mounted on vehicles.
As shown in Figure~\ref{fig:teaser} (middle), object scales vary mainly with distance.
The full data splits of WOD are too large for benchmarking methods.
Thus, we extracted 10\% size subsets from the predefined training split (798 sequences) and validation split (202 sequences)~\cite{WaymoOpenDataset_CVPR2020}.
Specifically,
we extracted splits based on the ones place of the frame index (frames 0, 10, ..., 190) in each sequence.
We call the subsets \texttt{f0train} and \texttt{f0val} splits.
Each sequence in the splits contains $\sim$20 frames (20\,s, 1\Hz),
and each frame contains five images for five cameras.
We used three categories (vehicle, pedestrian, and cyclist) following the official \textit{ALL\_NS} setting~\cite{WaymoOpenDataset_2D_detection_leaderboard} used in WOD competitions.

For the third dataset, we adopted the \MangasAbbr~\cite{Manga109_Matsui_MTAP2017, Manga109_Aizawa_IEEEMM2020}.
\MangasAbbr contains artificial images of manga (Japanese comics) and annotations for four categories (body, face, frame, and text).
Many characteristics differ from those of natural images.
Most images are grayscale.
The objects are highly overlapped~\cite{Manga109_detection_Ogawa_2018}.
As shown in Figure~\ref{fig:teaser} (right), object scales vary unrestrictedly with viewpoints and page layouts.
Small objects differ greatly from downsampled versions of large objects
because small objects are drawn with simple lines and points.
For example, small faces look like a sign ($\because$).
This characteristic may ruin techniques developed mainly for natural images.
We carefully selected 68, 4, and 15 volumes for training, validation, and testing splits,
and we call them the \texttt{68train}, \texttt{4val}, and \texttt{15test}, respectively.

\subsection{Motivation of Training Protocols}
\label{sec:usb_training_motivation}

\begin{table}[t]
	\setlength{\tabcolsep}{0.95mm}
	\renewcommand\arraystretch{0.85}
	\begin{center}
		\scalebox{0.7}{\begin{tabular}{lccccc}
			\toprule
			\multirow{2}{*}{Protocol}    & \multirow{2}{*}{Fair} & Suitable for & Strong  & Selectable &    Comparable    \\
			                             &                       &  each model  & results & divisions  & across divisions \\ \midrule
			A) Standard (short) training &          \cm          &              &         &            &                  \\
			B) Lawless (no regulations)  &                       &     \cm      &   \cm   &            &                  \\
			C) Ours w/o compatibility    &          \cm          &     \cm      &   \cm   &    \cm     &                  \\
			D) Ours                      &          \cm          &     \cm      &   \cm   &    \cm     &       \cm        \\ \bottomrule
		\end{tabular}}
	\end{center}
	\vspace{-3mm}
	\caption{
		Comparison of training protocols.
	}
	\label{table:training_protocols_comparison}
\end{table}

We describe the motivation of our training protocols with Table~\ref{table:training_protocols_comparison},
which compares existing protocols (A and B) and novel protocols (C and D).
Protocol A is the current standard training protocol within 24 epochs,
popularized by successive detectors, Detectron~\cite{Detectron2018}, and MMDetection~\cite{MMDetection}.
This protocol is fair but not suitable for slowly convergent models (\eg, DETR~\cite{DETR_ECCV2020}).
Protocol B is lawless without any regulations.
Participants can train their models with arbitrary settings suitable for them,
even if they are unfair settings
(\eg, standard training for existing methods
and longer training for proposed ones).
Since object detectors can achieve high accuracy with long training schedules and strong data augmentation~\cite{SpineNet_CVPR2020, EfficientDet_arXiv, SimpleCopyPaste_CVPR2021},
participants can buy stronger results~\cite{GreenAI_CACM2020}.

Since both existing protocols A and B have advantages and disadvantages,
we considered novel protocols to bridge them.
We first defined multiple divisions for training epochs, inspired by weight classes in sports.
This Protocol C enables fair comparison in each division.
Participants can select divisions according to their purposes and resources.
However, we cannot compare models across divisions.
To resolve this,
we propose Protocol D by introducing backward compatibility like the Universal Serial Bus.
As described above,
\emph{our protocols introduce a completely different paradigm from existing limited or unfair protocols.}

The training protocols mainly target resource-intensive factors that can increase the required resources 10 times or more.
This decision improves fairness without obstructing novel methods and practical settings that researchers can adopt without many resources.
We do not adopt factors that have large overlaps with inference efficiency, which has been considered in many previous studies.

\subsection{Training Protocols}
\label{sec:usb_training}

For fair training, we propose the \textit{USB training protocols} shown in Table~\ref{table:USB_training}.
By analogy with the backward compatibility of the Universal Serial Bus\footnote{
	Higher protocols can adapt the data transfer rate to lower protocols.
},
USB training protocols emphasize compatibility between protocols.
Importantly,
\textit{participants should report results with not only higher protocols but also lower protocols.}
For example, when a participant trains a model for 150 epochs with standard hyperparameters,
it corresponds to USB 3.0.
The participant should also report the results of models trained for 24 and 73 epochs in a paper.
This reveals the effectiveness of the method by ablating the effect of long training.
The readers of the paper can judge whether the method is useful for standard epochs.
Since many people do not have access to extensive computational resources,
such information is important.

\begin{table}
	\begin{minipage}[c]{0.485\hsize}
	\setlength{\tabcolsep}{0.8mm}
	\renewcommand\arraystretch{0.85}
	\begin{center}
		\scalebox{0.62}{\begin{tabular}{lrcll}
			\toprule
			Protocol  & Max epoch & AHPO & Compatibility     & Example                                                              \\ \midrule
			USB 1.0   &        24 & \xm  & ---               & 2$\times$ schedule~\cite{Detectron2018, RethinkingImageNet_ICCV2019} \\
			USB 2.0   &        73 & \xm  & USB 1.0           & 6$\times$ schedule~\cite{RethinkingImageNet_ICCV2019}                \\
			USB 3.0   &       300 & \xm  & USB 1.0, 2.0      & EfficientDet-D6~\cite{EfficientDet_CVPR2020}                         \\
			USB 3.1   &       300 & \cm  & USB 1.0, 2.0, 3.0 & YOLOv4~\cite{YOLOv4_2020}                                            \\
			Freestyle &  $\infty$ & \cm  & ---               & EfficientDet-D7x~\cite{EfficientDet_arXiv}                           \\ \bottomrule
		\end{tabular}}
	\end{center}
	\vspace{-3mm}
	\caption{
		USB training protocols.
		AHPO: Aggressive hyperparameter optimization.
	}
	\label{table:USB_training}
\end{minipage}\hfill
\begin{minipage}[c]{0.495\hsize}
	\setlength{\tabcolsep}{1.0mm}
	\renewcommand\arraystretch{0.85}
	\begin{center}
		\scalebox{0.62}{\begin{tabular}{lrrl}
			\toprule
			Protocol     & Max reso. &       Typical scale & Reference                                                        \\ \midrule
			Standard USB & 1,066,667 & 1333$\times$\;\:800 & Popular in COCO~\cite{COCO_ECCV2014, MMDetection, Detectron2018} \\
			Mini USB     &   262,144 &  512$\times$\;\:512 & Popular in VOC~\cite{PASCALVOC_IJCV2015, SSD_ECCV2016}           \\
			Micro USB    &    50,176 &  224$\times$\;\:224 & Popular in ImageNet~\cite{ImageNet_IJCV2015, ResNet_CVPR2016}    \\
			Large USB    & 2,457,600 &    1920$\times$1280 & WOD front cameras~\cite{WaymoOpenDataset_CVPR2020}               \\
			Huge USB     & 7,526,400 &    3360$\times$2240 & WOD methods (\hspace{1sp}\cite{Waymo2d_1st_2020}, ours)          \\
			Freestyle    &  $\infty$ &   ---\:\;\;\;\;\;\; & ---                                                              \\ \bottomrule
		\end{tabular}}
	\end{center}
	\vspace{-3mm}
	\caption{
		USB evaluation protocols.
	}
	\label{table:USB_resolutions}
\end{minipage}
\end{table}

The number of maximum epochs for USB 1.0 is 24, following a popular setting in COCO~\cite{MMDetection, Detectron2018}.
We adopted 73 epochs for USB 2.0,
where models trained from scratch can catch up with those trained from ImageNet pre-trained models~\cite{RethinkingImageNet_ICCV2019}.
This serves as a guideline for comparison between models with and without pre-training,
although perfectly fair comparisons are impossible considering the large differences caused by pre-training~\cite{Shinya_ICCVW2019}.
We adopted 300 epochs for USB 3.x such that 
YOLOv4~\cite{YOLOv4_2020} and most EfficientDet models~\cite{EfficientDet_arXiv} correspond to this protocol.
Models trained for more than 300 epochs are regarded as Freestyle.
They are not suitable for benchmarking methods,
although they may push the empirical limits of detectors~\cite{EfficientDet_arXiv, DETR_ECCV2020}.
The correspondences between Tables~\ref{table:training_protocols_comparison} and \ref{table:USB_training} are as follows:
Protocol A corresponds to only USB 1.0;
Protocol B corresponds to only Freestyle;
Protocol C corresponds to all protocols (divisions) in Table~\ref{table:USB_training} without compatibility; and
Protocol D corresponds to all protocols (divisions) in Table~\ref{table:USB_training} with compatibility.

In addition to long training schedules, hyperparameter optimization is resource-intensive.
If authors of a paper fine-tune hyperparameters for their architecture,
other people without sufficient computational resources cannot compare methods fairly.
For hyperparameters that need to be tuned exponentially, such as learning rates and $1 - m$ where $m$ denotes momentum,
the minimum ratio of hyperparameter choices should be greater than or equal to $2$
(\eg, choices $\{0.1, 0.2, 0.4, 0.8, ...\}$, $\{0.1, 0.2, 0.5, 1.0, ...\}$, and $\{0.1, 0.3, 1.0, ...\}$).
For hyperparameters that need to be tuned linearly,
the number of choices should be less than or equal to $11$
(\eg, choices $\{0.0, 0.1, 0.2, ..., 1.0\}$).
When participants perform aggressive hyperparameter optimization (AHPO) by manual fine-tuning or automatic algorithms,
$0.1$ is added to their number of protocols.
They should report both results with and without AHPO.
To further improve fairness without sacrificing the protocols' simplicity,
we consider it a kind of AHPO to use data augmentation techniques that more than double the time per epoch.

For models trained with annotations other than 2D bounding boxes (\eg, segmentation, keypoint, caption, and point cloud),
$0.5$ is added to their number of protocols.
Participants should also report results without such annotations if possible for their algorithms.

For ease of comparison, we limit the pre-training datasets to the three datasets and ImageNet-1k (ILSVRC 1,000-class classification)~\cite{ImageNet_IJCV2015}.
Other datasets are welcome only when the results with and without additional datasets are reported.
Participants should describe how to use the datasets
(\eg, fine-tuning models on WOD and \MangasAbbr from COCO pre-trained models,
or training a single model jointly~\cite{UniversalObjectDetection_CVPR2019, UniDet_Zhou_CVPR2022} on the three datasets).

\subsection{Evaluation Protocols}
\label{sec:usb_evaluation}

For fair evaluation, we propose the \textit{USB evaluation protocols} shown in Table~\ref{table:USB_resolutions}.
By analogy with the size variations of the Universal Serial Bus connectors for various devices,
USB evaluation protocols have variations in test image scales for various devices and applications.

The maximum resolution for Standard USB follows the popular test scale of 1333$\times$800 in the COCO benchmark~\cite{MMDetection, Detectron2018}.
For Mini USB, we limit the resolution based on 512$\times$512.
This resolution is popular in the \Pascal VOC benchmark~\cite{PASCALVOC_IJCV2015, SSD_ECCV2016}, which contains small images and large objects.
It is also popular in real-time detectors~\cite{EfficientDet_CVPR2020, YOLOv4_2020}.
We adopted a further small-scale 224$\times$224 for Micro USB.
This resolution is popular in ImageNet classification~\cite{ImageNet_IJCV2015, ResNet_CVPR2016}.
Although small object detection is extremely difficult,
it is suitable for low-power devices.
Additionally, this protocol enables people to manage object detection tasks using one or few GPUs.
To cover larger test scales than Standard USB, we define Large USB and Huge USB based on WOD resolutions
(see \AppendixSection~\ref{sec:details_experiments} for the top methods).
Although larger inputs (regarded as Freestyle) may be preferable for accuracy,
excessively large inputs reduce the practicality of detectors.

In addition to test image scales,
the presence and degree of Test-Time Augmentation (TTA) make large differences in accuracy and inference time.
When using TTA,
participants should report its details (including scales of multi-scale testing)
and results without TTA.

\subsection{Evaluation Metrics}

We mainly use the COCO metrics~\cite{COCO_ECCV2014, cocoapi} to evaluate the performance of detectors on each dataset.
We provide data format converters for WOD\footnote{\WaymoCOCO} and \MangasAbbr\footnote{\mangaapi}.
The COCO-style AP (CAP) for a dataset $d$ is calculated as
$\mathrm{CAP}_d = \frac{1}{|T|}\sum_{t \in T} \frac{1}{|C_d|}\sum_{c \in C_d} \mathrm{AP}_{t, c}$,
where
$T = \{0.5, 0.55, ..., 0.95\}$ denotes the predefined $10$ IoU thresholds,
$C_d$ denotes categories in the dataset $d$,
and $\mathrm{AP}_{t, c}$ denotes Average Precision (AP) for an IoU threshold $t$ and a category $c$.
For detailed analysis, five additional AP metrics (averaged over categories) are evaluated.
AP$_{50}$ and AP$_{75}$ denote AP at single IoU thresholds of $0.5$ and $0.75$, respectively.
\APS, \APM, and \APL are variants of CAP, where target objects are limited to
small (area $\leq 32^2$),
medium ($32^2 \leq$ area $\leq 96^2$),
and large ($96^2 \leq$ area) objects, respectively.
The area is measured using mask annotations for COCO and bounding box annotations for WOD and \MangasAbbr.

As the primary metric for USB,
we use the mean COCO-style AP (mCAP)
averaged over all datasets $D$ as
$\mathrm{mCAP} = \frac{1}{|D|}\sum_{d \in D} \mathrm{CAP}_d$.
Since USB adopts the three datasets described in Sec.~\ref{sec:usb_datasets},
$\mathrm{mCAP} = (\mathrm{CAP_{COCO}}+\mathrm{CAP_{WOD}}+\mathrm{CAP_{\MangasAbbr}}) / 3.$
Similarly,
we define five metrics from
AP$_{50}$, AP$_{75}$, \APS, \APM, and \APL
by averaging them over the datasets.

The three COCO-style scale-wise metrics (AP$_\textit{S}$, AP$_\textit{M}$, and AP$_\textit{L}$) are too coarse for detailed scale-wise analysis.
They confuse objects of significantly different scales.
For example, the absolute scale of a large object might be 100 or 1600.
Thus, we introduce finer scale-wise metrics.
We define the \textit{Absolute Scale AP (ASAP)} and \textit{Relative Scale AP (RSAP)}
using exponential thresholds.
ASAP partitions object scales based on absolute scales
$(0, 8, 16, 32, ..., 1024, \infty)$,
while RSAP partitions object scales based on relative scales
$(0, \frac{1}{256}, \frac{1}{128}, ..., \frac{1}{2}, 1)$.
We call the partitions by their maximum scales.

For ease of quantitative evaluation,
we limit the number of detections per image to 100 across all categories~\cite{cocoapi}.
For qualitative evaluation, participants may raise the limit to 300
because 1\% of images in the \MangasAbbr \texttt{15test} set contain more than 100 annotations.

\section{Experiments} \label{sec:experiments}

Here, we present benchmark results and analysis on USB.
See \AppendixSection~\ref{sec:details_experiments} for the details of the experimental settings and results,
including additional analysis and ablation studies.

\subsection{Experimental Settings} \label{sec:experimental_settings}

We compared and analyzed 15 methods.
With the ResNet-50-B~\cite{ResNet_CVPR2016, BagOfTricks_Classification_CVPR2019} backbone, we compared popular baseline methods:
(1) Faster R-CNN~\cite{Faster_R-CNN_NIPS2015} with FPN~\cite{FPN_CVPR2017}, (2) Cascade R-CNN~\cite{Cascade_R-CNN_CVPR2018},
(3) RetinaNet~\cite{RetinaNet_ICCV2017}, (4) ATSS~\cite{ATSS_CVPR2020}, (5) GFL~\cite{GFL_NeurIPS2020},
(6) DETR~\cite{DETR_ECCV2020}, (7) Deformable DETR~\cite{DeformableDETR_ICLR2021}, and (8) Sparse R-CNN~\cite{Sparse_R-CNN_CVPR2021}.
With ATSS~\cite{ATSS_CVPR2020}, we compared recent representative backbones and necks:
(9) Swin-T~\cite{SwinTransformer_ICCV2021}, (10) ConvNeXt-T~\cite{ConvNeXt_CVPR2022},
(11) SEPC without iBN~\cite{SEPC_CVPR2020}, and (12) DyHead~\cite{DyHead_CVPR2021}.
For a strong baseline, we trained (13) YOLOX-L~\cite{YOLOX_2021}, which adopts strong data augmentation.
We designed two additional detectors for USOD by collecting methods for multi-scale object detection.
(14) \OurOrig: ATSS~\cite{ATSS_CVPR2020} with SEPC (without iBN)~\cite{SEPC_CVPR2020}, Res2Net-50-v1b~\cite{Res2Net_TPAMI2020}, Deformable Convolutional Networks (DCN)~\cite{DCN_ICCV2017}, and multi-scale training.
(15) \OurAugust: A variant of \OurOrig designed around August 2020
with GFL~\cite{GFL_NeurIPS2020}, SyncBN~\cite{MegDet_CVPR2018}, iBN~\cite{SEPC_CVPR2020}, and the light use of DCN~\cite{DCN_ICCV2017, SEPC_CVPR2020}.
See \AppendixSection~\ref{sec:details_univ} for the details of the methods and architectures used in \Univs.

\vspace{-3mm}
\begin{table}[ht]
	\setlength{\tabcolsep}{0.9mm}
	\renewcommand\arraystretch{0.85}
	\begin{center}
	\begin{minipage}{0.53\hsize}
		\scalebox{0.62}{\begin{tabular}{lccc}
			\toprule
			Hyperparameters                          &      COCO       &       WOD       &   \MangasAbbr   \\ \midrule
			Learning rate for multi-stage detectors  &      0.02       &      0.02       &      0.16       \\
			Learning rate for single-stage detectors &      0.01       &      0.01       &      0.08       \\
			Test scale                               & 1333$\times$800 & 1248$\times$832 & 1216$\times$864 \\ \bottomrule
		\end{tabular}}
	\end{minipage}
	\begin{minipage}{0.18\hsize}
		\scalebox{0.62}{\begin{tabular}{lc}
				\toprule
				Hyperparam.  &  Common   \\ \midrule
				Epoch        &    12     \\
				Batch size   &    16     \\
				Momentum     &    0.9    \\
				Weight decay & $10^{-4}$ \\ \bottomrule
		\end{tabular}}
	\end{minipage}
	\end{center}
	\vspace{-3mm}
	\caption{
		Default hyperparameters.
		See \AppendixSection~\ref{sec:details_experiments} for exceptions.
	}
	\label{table:hyperparameters}
\end{table}

Our code is built on MMDetection~\cite{MMDetection}.
We trained models with Stochastic Gradient Descent (SGD) or AdamW~\cite{AdamW_ICLR2019}.
COCO models other than YOLOX~\cite{YOLOX_2021} were fine-tuned from ImageNet~\cite{ImageNet_IJCV2015} pre-trained backbones.
We trained the models for WOD and \MangasAbbr from the corresponding COCO pre-trained models
(some COCO models from MMDetection~\cite{MMDetection}).
The default hyperparameters are listed in Table~\ref{table:hyperparameters}.
Test scales were determined within the Standard USB protocol,
considering the typical aspect ratio of the images in each dataset.

\begin{table}
	\begin{minipage}[c]{\hsize}
	\centering
	\includegraphics[width=0.75\linewidth]{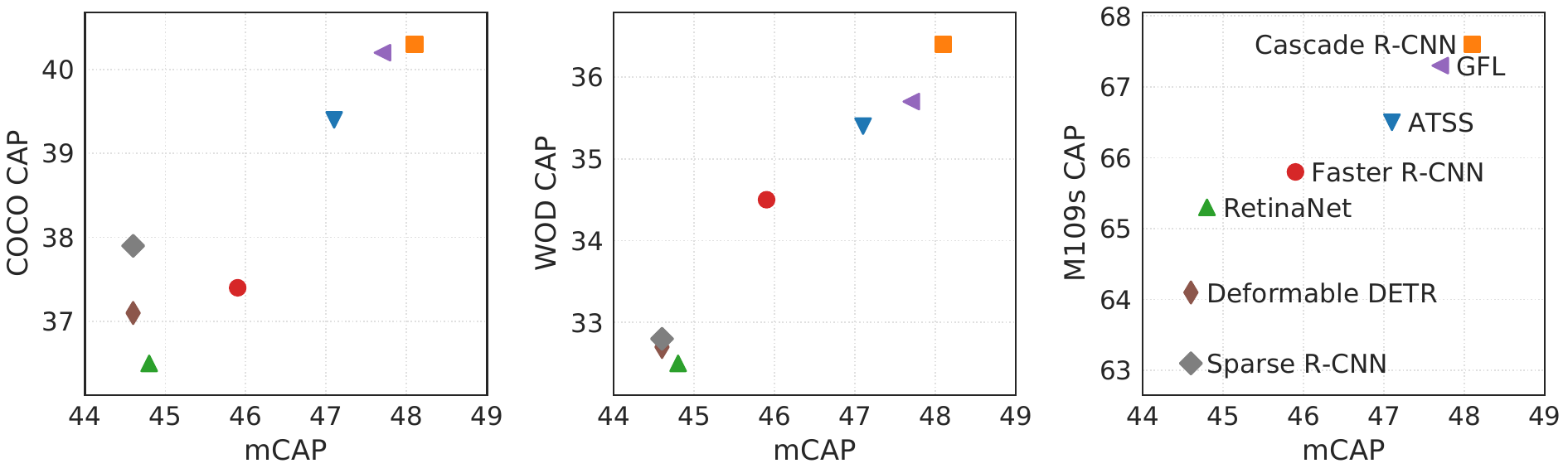}%
	\captionof{figure}{
		Correlation between mCAP and CAP on each dataset.
	}
	\label{fig:usb_correlation}
	\vspace{1mm}
	\end{minipage}
	\begin{minipage}[c]{0.55\hsize}
	\setlength{\tabcolsep}{0.8mm}
	\renewcommand\arraystretch{0.85}
	\begin{center}
		\scalebox{0.63}{\begin{tabularx}{1.58730\linewidth}{lc*{5}{>{\centering\arraybackslash}X}ccc}
			\toprule
			Method                                      &   mCAP    & AP$_{50}$ & AP$_{75}$ & \APS      & \APM      & \APL      &   COCO    &    WOD    & \MangasAbbr \\ \midrule
			Faster R-CNN~\cite{Faster_R-CNN_NIPS2015}   &   45.9    & 68.2      & 49.1      & 15.2      & 38.9      & 62.5      &   37.4    &   34.5    &    65.8     \\
			Cascade R-CNN~\cite{Cascade_R-CNN_CVPR2018} & \TB{48.1} & \TB{68.5} & \TB{51.5} & 15.6      & \TB{41.3} & \TB{65.9} & \TB{40.3} & \TB{36.4} &  \TB{67.6}  \\
			RetinaNet~\cite{RetinaNet_ICCV2017}         &   44.8    & 66.0      & 47.4      & 12.9      & 37.3      & 62.6      &   36.5    &   32.5    &    65.3     \\
			ATSS~\cite{ATSS_CVPR2020}                   &   47.1    & 68.0      & 50.2      & 15.5      & 39.5      & 64.7      &   39.4    &   35.4    &    66.5     \\
			GFL~\cite{GFL_NeurIPS2020}                  &   47.7    & 68.3      & 50.6      & \TB{15.8} & 39.9      & 65.8      &   40.2    &   35.7    &    67.3     \\
			DETR~\cite{DETR_ECCV2020}                   &   23.7    & 45.9      & 21.6      & 2.8       & 13.8      & 42.1      &   22.2    &   17.8    &    31.2     \\
			Deform. DETR~\cite{DeformableDETR_ICLR2021} &   44.6    & 67.0      & 47.3      & 13.8      & 36.1      & 62.6      &   37.1    &   32.7    &    64.1     \\
			Sparse R-CNN~\cite{Sparse_R-CNN_CVPR2021}   &   44.6    & 65.4      & 46.9      & 14.4      & 35.8      & 63.0      &   37.9    &   32.8    &    63.1     \\ \bottomrule
		\end{tabularx}}
	\end{center}
	\vspace{-3mm}
	\caption{
		Results of popular baseline methods.
	}
	\label{table:usb}
	\vspace{-6mm}
	\setlength{\tabcolsep}{0.8mm}
	\renewcommand\arraystretch{0.85}
	\begin{center}
		\scalebox{0.63}{\begin{tabularx}{1.58730\linewidth}{lc*{5}{>{\centering\arraybackslash}X}ccc}
			\toprule
			Backbone                                               &   mCAP    & AP$_{50}$ & AP$_{75}$ & \APS      & \APM      & \APL      &   COCO    &    WOD    & \MangasAbbr \\ \midrule
			ResNet-50-B~\cite{BagOfTricks_Classification_CVPR2019} &   47.1    & 68.0      & 50.2      & 15.5      & 39.5      & 64.7      &   39.4    &   35.4    &    66.5     \\
			Swin-T~\cite{SwinTransformer_ICCV2021}                 &   49.0    & 70.6      & 52.0      & 17.2      & 41.8      & 67.2      &   43.7    &   37.2    &    66.2     \\
			ConvNeXt-T~\cite{ConvNeXt_CVPR2022}\hspace{10pt}       & \TB{50.4} & \TB{71.8} & \TB{53.7} & \TB{17.3} & \TB{43.0} & \TB{69.0} & \TB{45.5} & \TB{38.3} &  \TB{67.4}  \\ \bottomrule
		\end{tabularx}}
	\end{center}
	\vspace{-3mm}
	\caption{
		ATSS~\cite{ATSS_CVPR2020} with different backbones.
	}
	\label{table:usb_backbone}
	\vspace{-6mm}
	\setlength{\tabcolsep}{0.8mm}
	\renewcommand\arraystretch{0.85}
	\begin{center}
		\scalebox{0.63}{\begin{tabularx}{1.58730\linewidth}{lc*{5}{>{\centering\arraybackslash}X}ccc}
			\toprule
			Neck                                            &   mCAP    & AP$_{50}$ & AP$_{75}$ & \APS      & \APM      & \APL      &   COCO    &    WOD    & \MangasAbbr \\ \midrule
			FPN~\cite{FPN_CVPR2017}                         &   47.1    & 68.0      & 50.2      & 15.5      & 39.5      & 64.7      &   39.4    &   35.4    &    66.5     \\
			FPN$+$SEPC~\cite{SEPC_CVPR2020}                 &   48.1    & 68.5      & 51.2      & 15.5      & 40.5      & 66.8      &   42.1    &   35.0    &    67.1     \\
			FPN$+$DyHead~\cite{DyHead_CVPR2021}\hspace{2pt} & \TB{49.4} & \TB{69.8} & \TB{52.9} & \TB{16.8} & \TB{43.0} & \TB{67.8} & \TB{43.3} & \TB{37.1} &  \TB{67.9}  \\ \bottomrule
		\end{tabularx}}
	\end{center}
	\vspace{-3mm}
	\caption{
		ATSS~\cite{ATSS_CVPR2020} with different necks.
	}
	\label{table:usb_neck}
	\vspace{-6mm}
	\setlength{\tabcolsep}{0.8mm}
	\renewcommand\arraystretch{0.85}
	\begin{center}
		\scalebox{0.63}{\begin{tabularx}{1.58730\linewidth}{lc*{5}{>{\centering\arraybackslash}X}ccc}
			\toprule
			Method                    &   mCAP    & AP$_{50}$ & AP$_{75}$ & \APS      & \APM      & \APL      &   COCO    &    WOD    & \MangasAbbr \\ \midrule
			YOLOX-L~\cite{YOLOX_2021} &   51.0    & 72.6      & 54.7      & \TB{21.2} & \TB{45.9} & 65.0      &   41.1    & \TB{41.6} &  \TB{70.2}  \\
			\OurOrig                  &   51.4    & 72.1      & 55.1      & 18.4      & 45.0      & 70.7      &   46.7    &   38.6    &    68.9     \\
			\OurAugust\hspace{2pt}    & \TB{52.1} & \TB{72.9} & \TB{55.5} & 19.2      & 45.8      & \TB{70.8} & \TB{47.5} &   39.0    &    69.9     \\ \bottomrule
		\end{tabularx}}
	\end{center}
	\vspace{-3mm}
	\caption{
		Results of strong baseline methods.
	}
	\label{table:usb_strong}
\end{minipage}\hfill
\begin{minipage}[c]{0.44\hsize}
	\centering
	\includegraphics[width=1.0\linewidth]{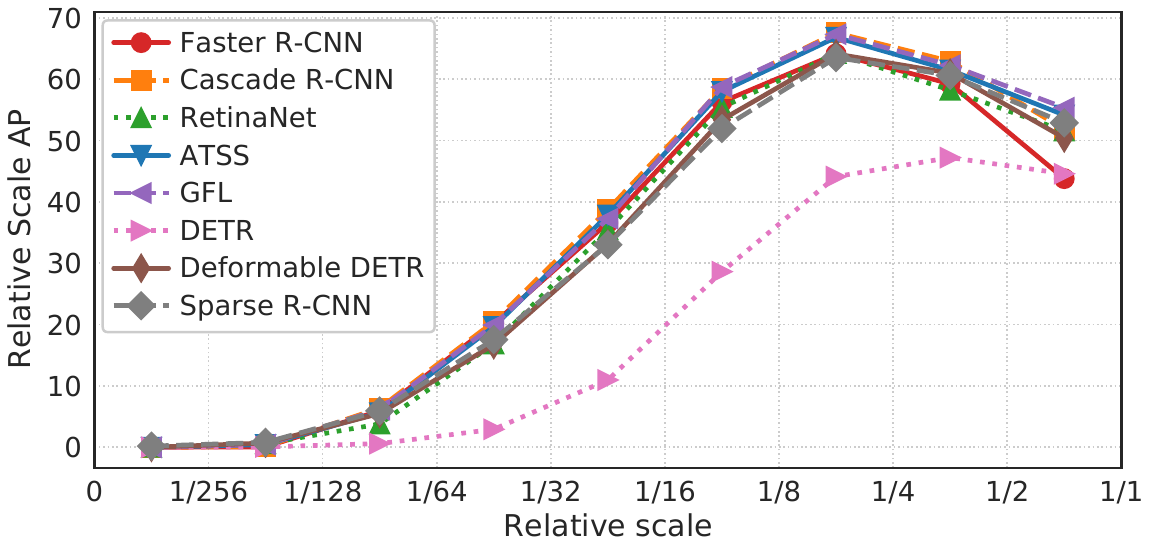}
	\captionof{figure}{
		Relative Scale AP of popular baseline methods.
	}
	\label{fig:usb_rsap}
	\vspace{3mm}
	\includegraphics[width=1.0\linewidth]{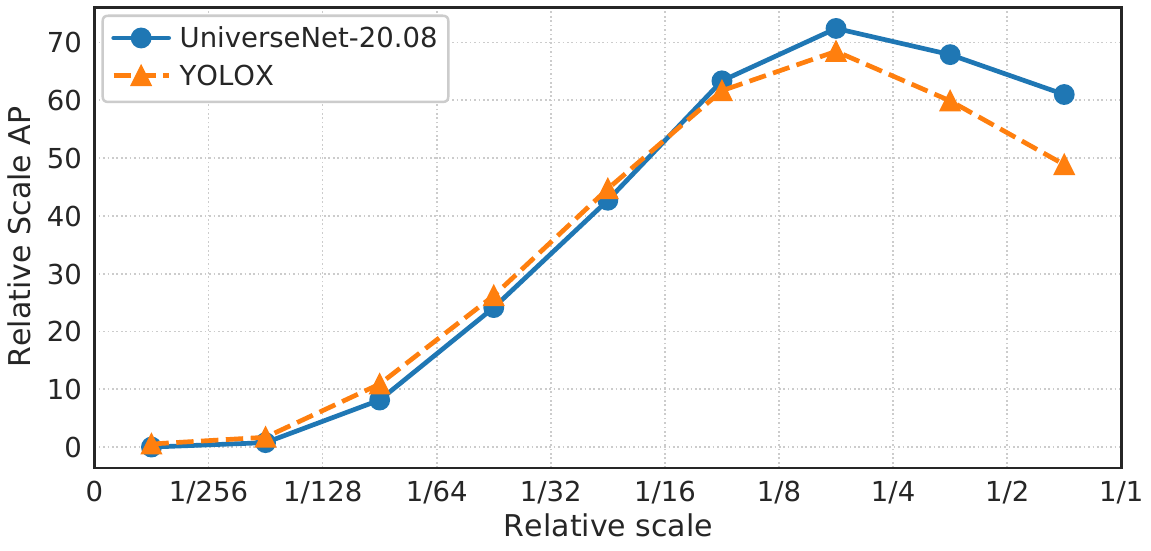}
	\captionof{figure}{
		Relative Scale AP of strong baseline methods.
	}
	\label{fig:usb_rsap_strong}

\end{minipage}

\end{table}

\subsection{Benchmark Results on USB}
\label{sec:usb_results}

\noindent
\textbf{Main results.}
We trained and evaluated the eight popular methods on USB.
All the methods follow the Standard USB 1.0 protocol.
The results are shown in Table~\ref{table:usb}.
Cascade R-CNN~\cite{Cascade_R-CNN_CVPR2018} achieves the highest results in almost all metrics.
The accuracy of DETR~\cite{DETR_ECCV2020} is low by a large margin.
We show the correlation between mCAP and CAP on each dataset in Figure~\ref{fig:usb_correlation}.
Faster R-CNN~\cite{Faster_R-CNN_NIPS2015} is underestimated on COCO.
Although Sparse R-CNN~\cite{Sparse_R-CNN_CVPR2021} is much more accurate than RetinaNet~\cite{RetinaNet_ICCV2017} on COCO, this is not true on the other datasets.
These results show the limitation of benchmarking with COCO only.

\noindent
\textbf{Backbones and necks.}
Tables~\ref{table:usb_backbone} and \ref{table:usb_neck} show the comparison results of the backbones and necks, respectively.
Swin-T~\cite{SwinTransformer_ICCV2021} shows lower AP than ResNet-50-B~\cite{ResNet_CVPR2016, BagOfTricks_Classification_CVPR2019} on \MangasAbbr.
SEPC~\cite{SEPC_CVPR2020} deteriorates WOD CAP.

\noindent
\textbf{Strong baselines.}
Table~\ref{table:usb_strong} shows the results of the three strong baselines.
\OurAugust achieves the highest mCAP of 52.1\%.
YOLOX-L~\cite{YOLOX_2021} shows better results on WOD and \MangasAbbr, which contain many small objects, possibly due to better \APS.

\noindent
\textbf{Scale-wise AP.}
We show RSAP on USB in Figures~\ref{fig:usb_rsap} and \ref{fig:usb_rsap_strong}.
Since the proposed metrics partition object scales evenly-spaced exponentially, we can confirm the continuous change.
RSAP does not increase monotonically but rather decreases at relative scales greater than $1/4$.
We cannot find this weakness from the coarse COCO-style scale-wise AP in Table~\ref{table:usb} \etc.
The difficulty of very large objects may be caused by truncation or unusual viewpoints~\cite{DiagDet_ECCV2012}.
The results also show that different methods are good at different scales.
We need further analysis in future research to develop methods that can detect both tiny and large objects.

\noindent
\textbf{Details on each dataset.}
We show detailed results on each dataset in \AppendixSection~\ref{sec:details_experiments}.
\APS on WOD is at most 12.0\%, which is much lower than \APS on COCO.
This highlights the limitation of COCO and current detectors.
Adding SEPC~\cite{SEPC_CVPR2020} to ATSS~\cite{ATSS_CVPR2020} decreases all metrics on WOD except for \APL.
We found that this reduction does not occur at large test scales in higher USB evaluation protocols.
Improvements by ATSS~\cite{ATSS_CVPR2020} on \MangasAbbr are smaller than those on COCO and WOD
due to the drop of face AP.
We conjecture that
this phenomenon comes from the domain differences discussed in Sec.~\ref{sec:usb_datasets} and prior work~\cite{Manga109_detection_Ogawa_2018}.

\noindent
\textbf{Qualitative results.}
We show some qualitative results of the best detector (\OurAugust) in Figure~\ref{fig:teaser}.
Although most detections are accurate,
it still suffers from classification error, localization error, and missing detections of tiny vehicles and small manga faces.

\section{Conclusions and Discussions}\label{sec:conclusions}

We introduced USB, a benchmark for universal-scale object detection.
To resolve unfair comparisons in existing benchmarks,
we established USB training/evaluation protocols.
With the benchmark,
we found weaknesses in existing methods to be addressed in future research.

There are several limitations to this work.
(1) USB has imbalances in domains and categories because it depends on the existing datasets that have large scale variations.
It will be an important direction to construct a well-balanced and more comprehensive benchmark that contains more domains and categories.
(2) The architectures and results of the 15 methods are still biased toward COCO
due to development and pre-training on COCO.
Less biased and more universal detectors should be developed in future research.
(3) We could not train detectors with higher protocols than USB 1.0 due to limited resources.
Although the compatibility enables comparison in low protocols,
still only well-funded researchers can compare detectors in high protocols.
Other efforts are also needed to ensure fairness and inclusion in research.
See \AppendixSection~\ref{sec:research_ethics} for discussion on other limitations and research ethics.

The current computer vision community places a high value on state-of-the-art results.
Thus, there is a large incentive to make unfair comparisons for overly accurate results, like DETR~\cite{DETR_ECCV2020} and EfficientDet~\cite{EfficientDet_CVPR2020}.
We need to create a system that emphasizes fair comparisons.
To improve effectiveness in broad areas, creating a checklist that can be incorporated into author/reviewer guidelines is a promising future direction.
We believe that our work is an important step toward realizing fair and inclusive research
by connecting various experimental settings.

	\paragraph*{Acknowledgments.}
	We are grateful to Dr. Hirokatsu Kataoka for helpful comments.
	We thank all contributors for the datasets and software libraries.
	The original image of Figure~\ref{fig:teaser} (left) is
	\href{https://www.flickr.com/photos/84326824@N00/428244913}{\textcolor{black}{\textit{satellite office}}} by Taiyo FUJII (\href{https://creativecommons.org/licenses/by/2.0/}{\textcolor{black}{CC BY 2.0}}).

\bibliography{universenet}

\clearpage
\appendix
\section*{Supplementary Material}

\section{Discussions on Research Ethics}
\label{sec:research_ethics}

\noindent
\textbf{Limitations.}
In addition to the limitations described in the main text, this work has the following limitations.
(1) USB depends on datasets with many instances.
Reliable scale-wise metrics for small datasets should be considered.
(2) USB does not cover the resolution of recent smartphone cameras (\eg, 4000$\times$3000).
Such high-resolution images may encourage completely different methods.
(3) USB, as well as UODB~\cite{UniversalObjectDetection_CVPR2019}, has a large imbalance in the number of images.
If participants train a unified detector~\cite{UniversalObjectDetection_CVPR2019, UniDet_Zhou_CVPR2022},
they will need strategies for dataset sampling~\cite{UniDet_Zhou_CVPR2022}.

\noindent
\textbf{Potential negative societal impacts.}
Improving the accuracy and universality of object detectors could improve the performance of autonomous weapons.
To mitigate the risk, we could develop more detectors for entertainment to increase people's happiness and decrease their hatred.
Besides, detectors might be misused for surveillance systems (\eg, as a part of person tracking methods).
To mitigate the risk, the computer vision community will need to have discussions with national and international organizations to regulate them appropriately.

\noindent
\textbf{Existing assets.}
We used the assets listed in Table~\ref{table:assets}.
See our codes for more details.
Refer to the papers~\cite{COCO_ECCV2014, WaymoOpenDataset_CVPR2020, Manga109_Aizawa_IEEEMM2020} and the URLs for how the datasets were collected.

\vspace{-3mm}
\begin{table*}[ht]
	\setlength{\tabcolsep}{1.0mm}
	\renewcommand\arraystretch{0.72}
	\begin{center}
		\scalebox{0.55}{\begin{tabular}{llll}
			\toprule
			Asset                                                      & Version    & URL                                                        & License                                          \\ \midrule
			COCO~\cite{COCO_ECCV2014}                                           & 2017       & \url{https://cocodataset.org/}                             & Annotations: CC-BY 4.0; images: various licenses \\
			WOD~\cite{WaymoOpenDataset_CVPR2020}                                & 1.2        & \url{https://waymo.com/open/}                              & Custom license                                   \\
			\Mangas~\cite{Manga109_Matsui_MTAP2017, Manga109_Aizawa_IEEEMM2020} & 2020.12.18 & \url{http://www.manga109.org/}                             & Custom license                                   \\ \midrule
			COCO API~\cite{cocoapi}                                             & 2.0        & \url{https://github.com/cocodataset/cocoapi}               & 2-Clause BSD License                             \\
			WOD (code)                                                          & ---        & \url{https://github.com/waymo-research/waymo-open-dataset} & Apache License 2.0                               \\
			Manga109 API                                                        & 0.3.1      & \url{https://github.com/manga109/manga109api}              & MIT License                                      \\
			MMDetection~\cite{MMDetection}                                      & 2.25.0     & \url{https://github.com/open-mmlab/mmdetection}            & Apache License 2.0                               \\ \bottomrule
		\end{tabular}}
	\end{center}
	\vspace{-3mm}
	\caption{
		Existing assets we used.
	}
	\label{table:assets}
\end{table*}
\vspace{-3mm}

\noindent
\textbf{Consent.}
See~\cite{Manga109_Aizawa_IEEEMM2020} for \MangasAbbr.
For the other datasets, we could not find whether and how consent was obtained.
It will be impossible to obtain consent from people recorded in datasets for autonomous driving such as WOD~\cite{WaymoOpenDataset_CVPR2020}.

\noindent
\textbf{Privacy.}
Faces and license plates in WOD~\cite{WaymoOpenDataset_CVPR2020} are blurred.
COCO images may harm privacy because they probably contain personally identifiable information.
However, COCO~\cite{COCO_ECCV2014} is so popular that the computer vision community cannot stop using it suddenly.
This paper will be a step toward reducing the dependence on COCO.

\noindent
\textbf{Offensive contents.}
\MangasAbbr covers various contents~\cite{Manga109_Aizawa_IEEEMM2020}.
This characteristic is useful to develop universal-scale object detectors.
One of the authors checked many images of the three datasets with eyes
and felt that some images in \MangasAbbr may be considered offensive
(\eg, violence in battle manga and nudity in romantic comedy).
Thus, researchers should be careful how they use it.
It is also valuable to develop methods to detect such scenes using the dataset.

\noindent
\textbf{Compute.}
Considering the numbers of training images,
training for USB takes about 1.7 ($\approx \frac{118287+79735+6467}{118287}$) times longer than that for COCO.
This is reasonable as a next-generation benchmark after COCO.
Furthermore, the proposed protocols provide incentives to avoid computationally intensive settings~\cite{EfficientDet_CVPR2020}.

\section{Details of Related Work}
\label{sec:details_related_work}

\subsection{Components for Multi-Scale Object Detection}

\noindent
\textbf{Backbones and modules.}
Inception module~\cite{Inception_CVPR2015} arranges $1{\times}1$, $3{\times}3$, and $5{\times}5$ convolutions to cover multi-scale regions.
Residual block~\cite{ResNet_CVPR2016} adds multi-scale features from shortcut connections and $3{\times}3$ convolutions.
ResNet-C and ResNet-D~\cite{BagOfTricks_Classification_CVPR2019}
replace the first layer of ResNet with the deep stem (three $3{\times}3$ convolutions)~\cite{Inceptionv3_CVPR2016}.
Res2Net module~\cite{Res2Net_TPAMI2020} stacks $3{\times}3$ convolutions hierarchically to represent multi-scale features.
Res2Net-v1b~\cite{Res2Net_TPAMI2020} adopts deep stem with Res2Net module.
Deformable convolution module in Deformable Convolutional Networks (DCN)~\cite{DCN_ICCV2017}
adjusts receptive field adaptively
by deforming the sampling locations of standard convolutions.
These modules are mainly used in backbones.

\noindent
\textbf{Necks.}
To combine and enhance backbones' representation, necks follow backbones.
Feature Pyramid Networks (FPN)~\cite{FPN_CVPR2017}
adopt top-down path and lateral connections like architectures for semantic segmentation.
Scale-Equalizing Pyramid Convolution (SEPC)~\cite{SEPC_CVPR2020}
introduces pyramid convolution across feature maps with different resolutions
and utilizes DCN to align the features.
Dynamic Head (DyHead)~\cite{DyHead_CVPR2021} improves SEPC with two types of attention mechanisms.

\noindent
\textbf{Heads and training sample selection.}
Faster R-CNN~\cite{Faster_R-CNN_NIPS2015} spreads multi-scale anchors over a feature map.
SSD~\cite{SSD_ECCV2016} spreads multi-scale anchors over multiple feature maps with different resolutions.
Adaptive Training Sample Selection (ATSS)~\cite{ATSS_CVPR2020} eliminates the need for multi-scale anchors
by dividing positive and negative samples according to object statistics across pyramid levels.

\noindent
\textbf{Multi-scale training and testing.}
Traditionally, the image pyramid is an essential technique to handle multi-scale objects~\cite{Rowley_PAMI1998}.
Although recent detectors can output multi-scale objects from a single-scale input,
many studies use multi-scale inputs to improve performance~\cite{Faster_R-CNN_NIPS2015, RetinaNet_ICCV2017, ATSS_CVPR2020, SEPC_CVPR2020}.
In a popular implementation~\cite{MMDetection},
multi-scale training randomly chooses a scale at each iteration for (training-time) data augmentation.
Multi-scale testing infers multi-scale inputs and merges their outputs for Test-Time Augmentation (TTA).
Scale Normalization for Image Pyramids (SNIP)~\cite{SNIP_Singh_CVPR2018} limits the range of object scales at each image scale during training and testing.

\subsection{Scale-Wise Metrics}

Many studies have introduced different scale-wise metrics~\cite{DiagDet_ECCV2012, Caltech_PAMI2012, COCO_ECCV2014, ImageNet_IJCV2015, TinyPerson_WACV2020, TIDE_ECCV2020, REVISE_ECCV2020}.
Unlike these studies, we introduce two types of finer scale-wise metrics based on the absolute scale and relative scale~\cite{TinyPerson_WACV2020}.
More importantly, we evaluated them on the datasets that have extensive scale variations and many instances in multiple domains.

\section{Details of Protocols}
\label{sec:details_protocols}

\subsection{Dataset Splits of \Mangas}

\begin{table}[t]
	\renewcommand\arraystretch{0.85}
	\begin{center}
		\scalebox{0.8}{\begin{tabular}{ll}
				\toprule
				Volume                    & Genre                                        \\ \midrule
				\multicolumn{2}{l}{\texttt{15test} \textit{set:}}                        \\
				Aku-Ham                   & Four-frame cartoons                          \\
				Bakuretsu! Kung Fu Girl   & Romantic comedy                              \\
				Doll Gun                  & Battle                                       \\
				Eva Lady                  & Science fiction                              \\
				Hinagiku Kenzan\textit{!} & Love romance                                 \\
				Kyokugen Cyclone          & Sports                                       \\
				Love Hina vol. 1          & Romantic comedy                              \\
				Momoyama Haikagura        & Historical drama                             \\
				Tennen Senshi G           & Humor                                        \\
				Uchi no Nyan's Diary      & Animal                                       \\
				Unbalance Tokyo           & Science fiction                              \\
				Yamato no Hane            & Sports                                       \\
				Youma Kourin              & Fantasy                                      \\
				Yume no Kayoiji           & Fantasy                                      \\
				Yumeiro Cooking           & Love romance                                 \\ \midrule
				\multicolumn{2}{l}{\texttt{4val} \textit{set:}}                          \\
				Healing Planet            & Science fiction                              \\
				Love Hina vol. 14         & Romantic comedy                              \\
				Seijinki Vulnus           & Battle                                       \\
				That's! Izumiko           & Fantasy                                      \\ \midrule
				\multicolumn{2}{l}{\texttt{68train} \textit{set: All the other volumes}} \\ \bottomrule
		\end{tabular}}
	\end{center}
	\vspace{-3mm}
	\caption{
		Manga109-s dataset splits (87 volumes in total).
	}
	\label{table:manga109_split}
\end{table}

The \textit{Manga109-s} dataset (87 volumes) is a subset of the full \textit{Manga109} dataset (109 volumes)~\cite{Manga109_Aizawa_IEEEMM2020}.
Unlike the full Manga109 dataset, the Manga109-s dataset can be used by commercial organizations.
The dataset splits for the full Manga109 dataset used in prior work~\cite{Manga109_detection_Ogawa_2018} cannot be used for the Manga109-s dataset.
We defined the Manga109-s dataset splits shown in Table~\ref{table:manga109_split}.
Unlike alphabetical order splits used in the prior work~\cite{Manga109_detection_Ogawa_2018}, we selected the volumes carefully.
The \texttt{15test} set was selected to be well-balanced for reliable evaluation.
Five volumes in the \texttt{15test} set were selected from the 10 test volumes used in~\cite{Manga109_detection_Ogawa_2018} to enable partially direct comparison.
All the authors of the \texttt{15test} and \texttt{4val} set are different from those of the \texttt{68train} set to evaluate generalizability.

\subsection{Number of Images}

There are
118,287 images in COCO \texttt{train2017},
5,000 in COCO \texttt{val2017},
79,735 in WOD \texttt{f0train},
20,190 in WOD \texttt{f0val},
6,467 in M109s \texttt{68train},
399 in M109s \texttt{4val}, and
1,289 in M109s \texttt{15test}.
Following prior work~\cite{Manga109_detection_Ogawa_2018},
we exclude M109s images without annotations because objects on irregular pages are not annotated.

We selected the test splits from images with publicly available annotations
to reduce labor for submissions.
Participants should not fine-tune hyperparameters based on the test splits
to prevent overfitting.

\subsection{Importance of Many Instances}

Here, we highlight the importance of a larger number of instances than UODB~\cite{UniversalObjectDetection_CVPR2019}.
We show that if we introduced scale-wise metrics to UODB, the results would be unreliable.
Watercolor2k, one of the datasets adopted by UODB, has $6$ classes and $27$ bicycle instances~\cite{CrossDomainDetection_Inoue_CVPR2018}.
If we equally divided the dataset for training and evaluation and they had the same number of small, medium, and large bicycles,
the average number of bicycles of a particular scale in the evaluation split would be $4.5$.
Since the $4.5$ bicycles affect $\frac{1}{6}$ of a scale-wise metric, a single error can change the results by $3.7$\%.
Thus, randomness can easily reverse the ranking between methods, making the benchmark results unreliable.

\subsection{Exceptions of Protocols}

The rounding error of epochs between epoch- and iteration-based training can be ignored when calculating the maximum epochs.
Small differences of eight pixels or less can be ignored when calculating the maximum resolutions.
For example, DSSD513~\cite{DSSD_2017} will be compared in Mini USB.

The number of additional images loaded for multi-image data augmentation techniques
(\eg, Between-Class Learning~\cite{BCL_CVPR2018}, mixup~\cite{mixup_ICLR2018}, RICAP~\cite{RICAP_TCSVT2020}, and Mosaic~\cite{YOLOv4_2020})
can be ignored when calculating the maximum epochs.
Even in that case, the time per epoch is considered according to another provision in \iftoggle{bmvcarxiv}{Sec.~\ref{sec:usb_training}.}{Sec.~3.5.}

\subsection{Constraints on Training Time}

We do not adopt constraints on training time as the major constraints of the training protocols because they have the following issues.

{\setlength{\leftmargini}{15pt}
	\begin{itemize}
		\setlength{\itemsep}{0.0mm}
		\setlength{\parskip}{0.0mm}
		\item It is difficult to measure training time on unified hardware.
		\item It is complicated to measure training time, calculate allowable epochs, and set  learning rate schedules for each model.
		\item It is difficult to compare with previous studies, which align the number of epochs.
		\item They will reduce the value of huge existing resources for standard training epochs (trained models, configuration files, and experimental results)
		provided by popular object detection libraries such as MMDetection~\cite{MMDetection}.
		\item They overemphasize implementation optimization rather than the trial and error of novel methods. 
		\item There are overlaps between the factors of training time and those of inference time.
	\end{itemize}
}
\noindent
The proposed constraints on training epochs are much easier to adopt and more reasonable.
Furthermore, our protocols compensate for the shortcomings of the epoch constraints by defining the provisions for hyperparameter optimization and data augmentation.

\subsection{Characteristics of Scale-Wise Metrics}
\label{sec:scale_metrics_characteristics}

ASAP and COCO-style scale-wise metrics are based on the absolute scale.
It has a weakness that it changes with image resizing.
To limit inference time and GPU memory consumption, and to ensure fair comparisons, input image scales are typically resized.
If they are smaller than the original image scales,
relative scales have direct effects on accuracy rather than absolute scales.
Furthermore, objects with the same absolute scale in the original images may have different absolute scales in the input images.
Fluctuating object scale thresholds is not desirable for scale-wise metrics.

In addition,
ASAP is not suitable for evaluating accuracy for very large objects.
It may be impossible to calculate ASAP for large absolute scales on some datasets.
In the case of USB, we cannot calculate ASAP$_\infty$ on COCO
because the absolute scales of COCO objects are smaller than 1024
(we filled ASAP$_\infty$ on COCO with zero in experiments).
Furthermore,
ASAP for large absolute scales may show unusual behavior.
For example, in the evaluation of ASAP$_\infty$ on \MangasAbbr,
all predictions larger than 1024 of absolute scales have larger IoUs than $0.5$ with an object of image resolution size (1654$\times$1170). 

We prefer RSAP to ASAP due to the above-mentioned weaknesses of ASAP.
Absolute scales may be important depending on whether and how participants resize images.
In that case, RSAP and ASAP can be used complementarily.

\begin{table*}[t]
	\setlength{\tabcolsep}{0.6mm}
	\renewcommand\arraystretch{0.85}
	\newcommand{\METHOD}{\multirow{2}{*}[-0.5\dimexpr \aboverulesep + \belowrulesep + \cmidrulewidth]{Method}}
	\newcommand{\FPS}{\multirow{2}{*}[-0.5\dimexpr \aboverulesep + \belowrulesep + \cmidrulewidth]{FPS}}
	\newcommand{\Bb}{Backbone}
	\newcommand{\CMRs}{\cmidrule(l{.2em}r{.2em}){2-3}\cmidrule(l{.2em}r{.2em}){4-6}\cmidrule(l{.2em}r{.2em}){7-9}\cmidrule(l{.2em}r{.2em}){10-10}\cmidrule{12-17}}
	\newcommand{\CITERRB}{\cite{Res2Net_TPAMI2020}}
	\newcommand{\CITEiBNSBN}{\cite{SEPC_CVPR2020, MegDet_CVPR2018}}
	\begin{center}
		\scalebox{0.615}{\begin{tabular}{lcccccccccc@{\hspace{.9em}}cccccc}
				\toprule
				\METHOD                                      & \multicolumn{2}{c}{Head} & \multicolumn{3}{c}{Neck} & \multicolumn{3}{c}{\Bb} &  Input  & \FPS &             \multicolumn{6}{c}{COCO (1$\times$ schedule)}             \\
				\CMRs                                        & ATSS &        GFL         & PConv  &  DCN  &    iBN     & Res2 &  DCN  &   SyncBN    & MStrain &      &    AP     & AP$_{50}$ & AP$_{75}$ &   \APS    &   \APM    &   \APL    \\ \midrule
				RetinaNet~\cite{RetinaNet_ICCV2017} &      &                    &     &       &            &      &       &          &         & 33.9 &   36.5    &   55.4    &   39.1    &   20.4    &   40.3    &   48.1    \\
				ATSS~\cite{ATSS_CVPR2020}           & \cm  &                    &     &       &            &      &       &          &         & 35.2 &   39.4    &   57.6    &   42.8    &   23.6    &   42.9    &   50.3    \\
				GFL~\cite{GFL_NeurIPS2020}      & \cm  &        \cm         &     &       &            &      &       &          &         & 37.2 &   40.2    &   58.4    &   43.3    &   23.3    &   44.0    &   52.2    \\
				\ATSEPC~\cite{ATSS_CVPR2020, SEPC_CVPR2020}  & \cm  &                    & \cm & P, LC &            &      &       &          &         & 25.0 &   42.1    &   59.9    &   45.5    &   24.6    &   46.1    &   55.0    \\
				\OurOrig                                     & \cm  &                    & \cm & P, LC &            & \cm  & c3-c5 &          &   \cm   & 17.3 &   46.7    &   65.0    &   50.7    &   29.2    &   50.6    &   61.4    \\
				\OurGFL                                      & \cm  &        \cm         & \cm & P, LC &            & \cm  & c3-c5 &          &   \cm   & 17.5 &   47.5    &   65.8    &   51.8    &   29.2    &   51.6    &   62.5    \\
				\OurAugustD                                  & \cm  &        \cm         & \cm & P, LC &    \cm     & \cm  & c3-c5 &   \cm    &   \cm   & 17.3 & \TB{48.6} & \TB{67.1} & \TB{52.7} & \TB{30.1} & \TB{53.0} & \TB{63.8} \\
				\OurAugust                                   & \cm  &        \cm         & \cm &  LC   &    \cm     & \cm  &  c5   &   \cm    &   \cm   & 24.9 &   47.5    &   66.0    &   51.9    &   28.9    &   52.1    &   61.9    \\ \midrule
				\OurAugust w/o SEPC~\cite{SEPC_CVPR2020}                 & \cm  &        \cm         &     &       &            & \cm  &  c5   &   \cm    &   \cm   & 26.7 &   45.8    &   64.6    &   50.0    &   27.6    &   50.4    &   59.7    \\
				\OurAugust w/o Res2Net-v1b~\CITERRB                      & \cm  &        \cm         & \cm &  LC   &    \cm     &      &  c5   &   \cm    &   \cm   & 32.8 &   44.7    &   62.8    &   48.4    &   27.1    &   48.8    &   59.5    \\
				\OurAugust w/o DCN~\cite{DCN_ICCV2017}                   & \cm  &        \cm         & \cm &       &    \cm     & \cm  &       &   \cm    &   \cm   & 27.8 &   45.9    &   64.5    &   49.8    &   28.9    &   49.9    &   59.0    \\
				\OurAugust w/o iBN, SyncBN~\CITEiBNSBN                   & \cm  &        \cm         & \cm &  LC   &            & \cm  &  c5   &          &   \cm   & 25.7 &   45.8    &   64.0    &   50.2    &   27.9    &   50.0    &   59.8    \\
				\OurAugust w/o MStrain                                   & \cm  &        \cm         & \cm &  LC   &    \cm     & \cm  &  c5   &   \cm    &         & 24.8 &   45.9    &   64.5    &   49.6    &   27.4    &   50.5    &   60.1    \\ \bottomrule
		\end{tabular}}
	\end{center}
	\vspace{-2.5mm}
	\caption{
		Architectures of \Univs with a summary of ablation studies on COCO \texttt{minival}.
		See Sec.~\ref{sec:coco_ablation} for step-by-step improvements.
		All results are based on MMDetection~\cite{MMDetection} v2.
		The ``Head'' methods (ATSS and GFL) affect losses and training sample selection.
		Res2: Res2Net-v1b~\cite{Res2Net_TPAMI2020}.
		PConv (Pyramid Convolution) and iBN (integrated Batch Normalization) are the components of SEPC~\cite{SEPC_CVPR2020}.
		The DCN columns indicate where to apply DCN.
		``P'': The PConv modules in the combined head of SEPC~\cite{SEPC_CVPR2020}.
		``LC'': The extra head of SEPC for localization and classification~\cite{SEPC_CVPR2020}.
		``c3-c5'': conv3\_x, conv4\_x, and conv5\_x layers in ResNet-style backbones~\cite{ResNet_CVPR2016}.
		``c5'': conv5\_x layers in ResNet-style backbones~\cite{ResNet_CVPR2016}.
		\ATSEPC: ATSS with SEPC (without iBN).
		MStrain: Multi-scale training.
		FPS: Frames per second on one V100 with mixed precision.
	}
	\label{table:coco_ablation_details}
\end{table*}

\section{Details of \Univs}
\label{sec:details_univ}

For fast and accurate detectors for USOD, we designed \Univs.
We adopted single-stage detectors for efficiency.
We show the detailed architectures in Table~\ref{table:coco_ablation_details}.

As a baseline model,
we used the RetinaNet~\cite{RetinaNet_ICCV2017} implemented in MMDetection~\cite{MMDetection}.
Specifically,
the backbone is ResNet-50-B~\cite{BagOfTricks_Classification_CVPR2019} (a variant of ResNet-50~\cite{ResNet_CVPR2016}, also known as the PyTorch style).
The neck is FPN~\cite{FPN_CVPR2017}.
We used focal loss~\cite{RetinaNet_ICCV2017}, single-scale training, and single-scale testing.

Built on the RetinaNet baseline,
we designed \textit{\OurOrig} by collecting human wisdom about multi-scale object detection as of May 2020.
We used ATSS~\cite{ATSS_CVPR2020} and SEPC without iBN~\cite{SEPC_CVPR2020}
(hereafter referred to as \textit{ATSEPC}).
The backbone is Res2Net-50-v1b~\cite{Res2Net_TPAMI2020}.
We adopted Deformable Convolutional Networks (DCN)~\cite{DCN_ICCV2017} in the backbone and neck.
We used multi-scale training.
Unless otherwise stated, we used single-scale testing for efficiency.

By adding GFL~\cite{GFL_NeurIPS2020}, SyncBN~\cite{MegDet_CVPR2018}, and iBN~\cite{SEPC_CVPR2020},
we designed three variants of \OurOrig
around August 2020.
\textit{\OurAugustD} heavily uses DCN~\cite{DCN_ICCV2017}.
\textit{\OurAugust} speeds up inference (and training) by the light use of DCN~\cite{DCN_ICCV2017, SEPC_CVPR2020}.
\textit{\OurAugustS} further speeds up inference using the ResNet-50-C~\cite{BagOfTricks_Classification_CVPR2019} backbone.

\section{Details of Experiments}
\label{sec:details_experiments}

Here, we show the details of experimental settings and results.
See also the code to reproduce our settings including minor hyperparameters.

\subsection{Common Settings}

We follow the learning rate schedules of MMDetection~\cite{MMDetection}, which are similar to those of Detectron~\cite{Detectron2018}.
Specifically, the learning rates are reduced by 10$\times$ in two predefined epochs.
Epochs for the first learning rate decay, the second decay, and ending training are
$(8, 11, 12)$ for the 1$\times$ schedule,
$(16, 22, 24)$ for the 2$\times$ schedule, and
$(16, 19, 20)$ for the 20e schedule.
To avoid overfitting by small learning rates~\cite{Shinya_ICCVW2019}, the 20e schedule is reasonable.
We mainly used the 1$\times$ schedule (12 epochs).

We mainly used ImageNet~\cite{ImageNet_IJCV2015} pre-trained backbones that are standard in MMDetection~\cite{MMDetection}.
Some pre-trained backbones not supported in MMDetection were downloaded from the repositories of Res2Net~\cite{Res2Net_TPAMI2020}
and Swin Transformer~\cite{SwinTransformer_ICCV2021}.
We used the COCO pre-trained models of the MMDetection~\cite{MMDetection} repository for several existing methods
(Faster R-CNN~\cite{Faster_R-CNN_NIPS2015} with FPN~\cite{FPN_CVPR2017}, Cascade R-CNN~\cite{Cascade_R-CNN_CVPR2018},
RetinaNet~\cite{RetinaNet_ICCV2017}, ATSS~\cite{ATSS_CVPR2020}, GFL~\cite{GFL_NeurIPS2020}, and Sparse R-CNN~\cite{Sparse_R-CNN_CVPR2021}).
We trained most models with mixed precision and 4 GPUs ($\times$ 4 images per GPU).
We mainly used NVIDIA T4 GPUs on the Google Cloud Platform.
All results on USB and all results of \Univs are single model results without ensemble.
We could not train each object detector multiple times with different random seeds to report error bars
because training object detectors is too computationally expensive.

\subsection{Settings for Specific Methods}

Many recent detectors~\cite{DETR_ECCV2020, SwinTransformer_ICCV2021} adopt the AdamW optimizer~\cite{AdamW_ICLR2019}.
Since AdamW does not necessarily give better results than SGD, we follow the optimizer settings of the official implementations and MMDetection~\cite{MMDetection}.
Specifically, for COCO and WOD, we used AdamW with an initial learning rate of $10^{-4}$ for
DETR~\cite{DETR_ECCV2020},
Deformable DETR~\cite{DeformableDETR_ICLR2021},
ATSS~\cite{ATSS_CVPR2020} with Swin-T~\cite{SwinTransformer_ICCV2021}, and
ATSS~\cite{ATSS_CVPR2020} with ConvNeXt-T~\cite{ConvNeXt_CVPR2022}, and
$2.5{\times}10^{-5}$ for Sparse R-CNN~\cite{Sparse_R-CNN_CVPR2021}.
The learning rate of the backbone is $10^{-5}$ for DETR~\cite{DETR_ECCV2020} and Deformable DETR~\cite{DeformableDETR_ICLR2021}.
For training ATSS~\cite{ATSS_CVPR2020} with ConvNeXt-T~\cite{ConvNeXt_CVPR2022},
we used layer-wise learning rate decay and a stochastic depth rate of 0.2 (see \cite{ConvNeXt_CVPR2022}).

We trained YOLOX-L~\cite{YOLOX_2021} with SGD and an initial learning rate of $2.5{\times}10^{-3}$ for COCO and WOD.
The test scale on COCO is 1024$\times$1024.
Since YOLOX models are trained from scratch~\cite{YOLOX_2021},
we trained a COCO model for 36 epochs (USB 2.0) such that it achieves similar AP to a model trained for 12 epochs from an ImageNet pre-trained model~\cite{Shinya_ICCVW2019}.
We also trained a COCO model for 24 epochs (USB 1.0).

We used multi-scale training for YOLOX-L~\cite{YOLOX_2021} and \Univs.
The range of shorter side pixels for most models is 480--960, following prior work~\cite{SEPC_CVPR2020}.
That for YOLOX-L~\cite{YOLOX_2021} on COCO is 512--1024, and
that for \Univs on WOD is 640--1280.
Since we do not have sufficient computational resources, these hyperparameters have room for improvement.

For comparison with state-of-the-art methods on COCO,
we used the 2$\times$ schedule (24 epochs) for most models
and the 20e schedule (20 epochs) for \OurAugustD due to overfitting with the 2$\times$ schedule.
For comparison with state-of-the-art methods on WOD,
we trained \OurOrig on the WOD full training set for 7 epochs.
We used a learning rate of $10^{-3}$ for 6 epochs and $10^{-4}$ for the last epoch.

For comparison with state-of-the-art methods with TTA on COCO,
we used soft voting with 13-scale testing and horizontal flipping following the original implementation of ATSS~\cite{ATSS_CVPR2020}.
Specifically,
shorter side pixels are (400, 500, 600, 640, 700, 800, 900, 1000, 1100, 1200, 1300, 1400, 1800),
while longer side pixels are their 1.667$\times$.
For the 13 test scales, target objects are limited to corresponding 13 predefined ranges
((96, $\infty$), (96, $\infty$), (64, $\infty$), (64, $\infty$), (64, $\infty$), (0, $\infty$), (0, $\infty$), (0, $\infty$), (0, 256), (0, 256), (0, 192), (0, 192), (0, 96)),
where each tuple denotes the minimum and maximum absolute scales.
We also evaluated 5-scale TTA
because the above-mentioned ATSS-style TTA is slow.
We picked (400, 600, 800, 1000, 1200) for shorter side pixels,
and ((96, $\infty$), (64, $\infty$), (0, $\infty$), (0, $\infty$), (0, 256)) for absolute scale ranges.

For \MangasAbbr, we used learning rates 8$\times$ those of COCO and WOD.
The value is roughly tuned based on a preliminary experiment with the RetinaNet~\cite{RetinaNet_ICCV2017} baseline model.
For training ATSS~\cite{ATSS_CVPR2020} with different backbones (Swin-T~\cite{SwinTransformer_ICCV2021} and ConvNeXt-T~\cite{ConvNeXt_CVPR2022}),
we used an initial learning rate of $4{\times}10^{-4}$,
roughly tuned from choices $\{2{\times}10^{-4}, 4{\times}10^{-4}, 8{\times}10^{-4}\}$.

\begin{table}[t]
\begin{minipage}[c]{0.44\hsize}
	\centering
	\includegraphics[width=1.0\linewidth]{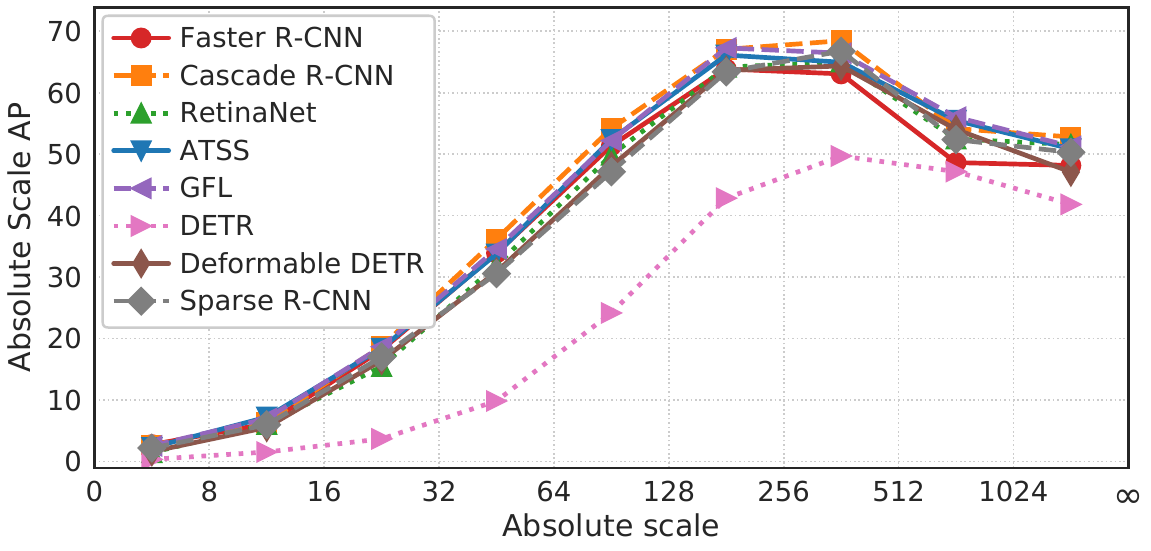}
	\vspace{-4.5mm}
	\captionof{figure}{
		Absolute Scale AP of popular baseline methods.
	}
	\label{fig:usb_asap}
\end{minipage}\hfill
\begin{minipage}[c]{0.44\hsize}
	\centering
	\includegraphics[width=1.0\linewidth]{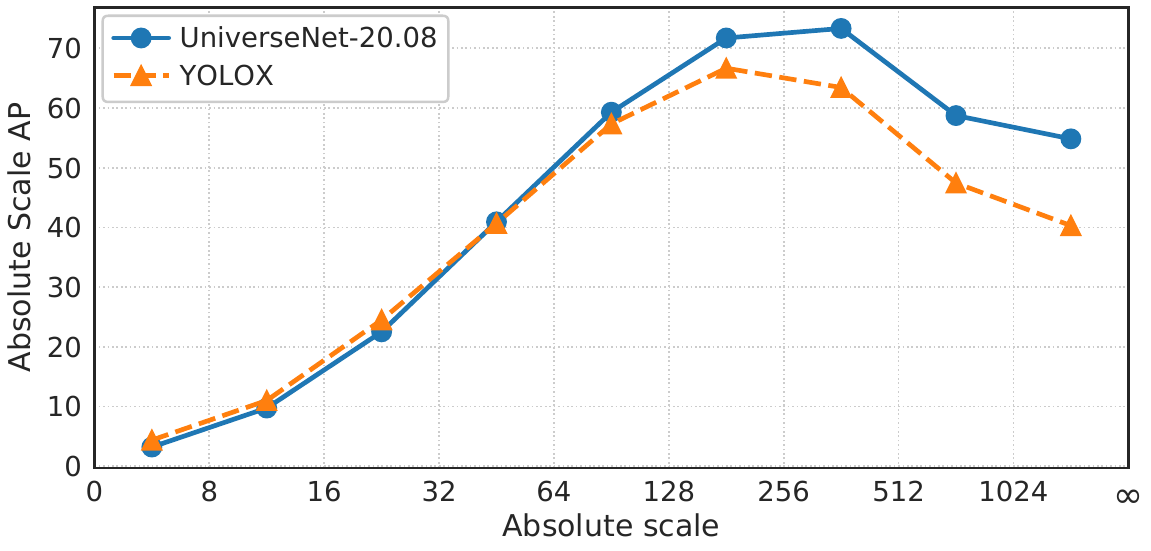}
	\vspace{-4.5mm}
	\captionof{figure}{
		Absolute Scale AP of strong baseline methods.
	}
	\label{fig:usb_asap_strong}
\end{minipage}
\end{table}

\subsection{Evaluation with Scale-Wise Metrics}

We show RSAP and ASAP of popular baseline methods on USB in
\iftoggle{bmvcarxiv}{Figures~\ref{fig:usb_rsap}}{Figures~4}
and \ref{fig:usb_asap}, respectively.
They do not increase monotonically but rather decrease at relative scales greater than $1/4$
or absolute scales greater than $512$.
The difficulty of very large objects may be caused by truncation or unusual viewpoints~\cite{DiagDet_ECCV2012}.
Except for the issues of ASAP$_\infty$ discussed in Sec.~\ref{sec:scale_metrics_characteristics},
ASAP shows similar changes to RSAP.
We conjecture that this is because image resolutions do not change much in each dataset of USB.
RSAP$_{\frac{1}{64}}$ and ASAP$_{16}$ are less than 10\%,
which indicates the difficulty of tiny object detection~\cite{TinyPerson_WACV2020}.
RetinaNet~\cite{RetinaNet_ICCV2017} shows low AP for small objects,
while Faster R-CNN~\cite{Faster_R-CNN_NIPS2015} with FPN~\cite{FPN_CVPR2017} shows low AP for large objects.
These results are consistent with the benchmark results of previous work~\cite{SpeedAccuracyTradeOffs_CVPR2017},
which compares SSD~\cite{SSD_ECCV2016} with Faster R-CNN without FPN on COCO.
For further analysis, it will be worth comparing the design choice of pyramid levels~\cite{RetinaNet_ICCV2017, TinyPerson_WACV2020}.

We show RSAP and ASAP of strong baseline methods on USB in
\iftoggle{bmvcarxiv}{Figures~\ref{fig:usb_rsap_strong}}{Figures~5}
and \ref{fig:usb_asap_strong}, respectively.
\OurAugust is more accurate for large objects, and YOLOX-L~\cite{YOLOX_2021} is more accurate for small objects.
We conjecture that the former is due to SEPC~\cite{SEPC_CVPR2020} and DCN~\cite{DCN_ICCV2017} (see Sec.~\ref{sec:coco_ablation} and \ref{sec:test_scales}).
For the latter, YOLOX may be biased toward small object detection
due to Mosaic augmentation~\cite{YOLOv4_2020} and
the absence of pre-training~\cite{Shinya_ICCVW2019} on ImageNet that has larger objects than COCO~\cite{SNIP_Singh_CVPR2018}.

\clearpage

{%
\renewcommand{\topfraction}{.99}
\renewcommand{\bottomfraction}{.99}
\renewcommand{\textfraction}{.01}

\subsection{Details on Each Dataset}

Tables~\ref{table:coco_minival}, \ref{table:waymo_f0_train_f0val_832}, and \ref{table:Manga109s_15test} show the results on COCO, WOD, and \MangasAbbr, respectively.

\begin{table}[hb]
	\begin{minipage}{\linewidth}
	\setlength{\tabcolsep}{0.15em}
	\renewcommand\arraystretch{0.85}
	\begin{center}
		\scalebox{0.7}{\begin{tabularx}{0.75\textwidth}{l*{6}{>{\centering\arraybackslash}X}}
			\toprule
			Method                                                             & AP        & AP$_{50}$ & AP$_{75}$ & \APS      & \APM      & \APL      \\ \midrule
			Faster R-CNN~\cite{Faster_R-CNN_NIPS2015}                          & 37.4      & 58.1      & 40.4      & 21.2      & 41.0      & 48.1      \\
			Cascade R-CNN~\cite{Cascade_R-CNN_CVPR2018}                        & \TB{40.3} & \TB{58.6} & \TB{44.0} & 22.5      & 43.8      & 52.9      \\
			RetinaNet~\cite{RetinaNet_ICCV2017}                                & 36.5      & 55.4      & 39.1      & 20.4      & 40.3      & 48.1      \\
			ATSS~\cite{ATSS_CVPR2020}                                          & 39.4      & 57.6      & 42.8      & \TB{23.6} & 42.9      & 50.3      \\
			GFL~\cite{GFL_NeurIPS2020}                                         & 40.2      & 58.4      & 43.3      & 23.3      & \TB{44.0} & 52.2      \\
			DETR~\cite{DETR_ECCV2020}                                          & 22.2      & 39.7      & 22.0      & 6.7       & 22.4      & 36.8      \\
			Deformable DETR~\cite{DeformableDETR_ICLR2021}                     & 37.1      & 55.9      & 39.9      & 19.3      & 40.7      & 50.5      \\
			Sparse R-CNN~\cite{Sparse_R-CNN_CVPR2021}                          & 37.9      & 56.0      & 40.5      & 20.7      & 40.0      & \TB{53.5} \\ \midrule
			ATSS~\cite{ATSS_CVPR2020}$+$Swin-T~\cite{SwinTransformer_ICCV2021} & 43.7      & 63.0      & 47.1      & \TB{28.2} & 47.0      & 56.8      \\
			ATSS~\cite{ATSS_CVPR2020}$+$ConvNeXt-T~\cite{ConvNeXt_CVPR2022}    & \TB{45.5} & \TB{64.7} & \TB{49.5} & 28.0      & \TB{49.1} & \TB{59.1} \\ \midrule
			ATSS~\cite{ATSS_CVPR2020}$+$SEPC~\cite{SEPC_CVPR2020}              & 42.1      & 59.9      & 45.5      & 24.6      & 46.1      & 55.0      \\
			ATSS~\cite{ATSS_CVPR2020}$+$DyHead~\cite{DyHead_CVPR2021}          & \TB{43.3} & \TB{60.9} & \TB{47.2} & \TB{26.6} & \TB{47.1} & \TB{55.8} \\ \midrule
			YOLOX-L~\cite{YOLOX_2021} (USB 1.0)                                & 37.8      & 56.5      & 41.5      & 26.4      & 42.0      & 41.1      \\
			YOLOX-L~\cite{YOLOX_2021} (USB 2.0)                                & 41.1      & 60.2      & 45.3      & 29.1      & 45.8      & 45.5      \\
			\OurOrig                                                           & 46.7      & 65.0      & 50.7      & \TB{29.2} & 50.6      & 61.4      \\
			\OurAugust w/o MStrain                                             & 45.9      & 64.5      & 49.6      & 27.4      & 50.5      & 60.1      \\
			\OurAugust                                                         & \TB{47.5} & \TB{66.0} & \TB{51.9} & 28.9      & \TB{52.1} & \TB{61.9} \\ \bottomrule
		\end{tabularx}}
	\end{center}
	\vspace{-3mm}
	\caption{
		Results on COCO \texttt{minival}.
	}
	\label{table:coco_minival}

	\setlength{\tabcolsep}{0.15em}
	\renewcommand\arraystretch{0.85}
	\begin{center}
		\scalebox{0.7}{\begin{tabularx}{0.95\textwidth}{l*{9}{>{\centering\arraybackslash}X}}
			\toprule
			Method                                                             & AP        & AP$_{50}$ & AP$_{75}$ & \APS      & \APM      & \APL      & veh.      & ped.      & cyc.      \\ \midrule
			Faster R-CNN~\cite{Faster_R-CNN_NIPS2015}                          & 34.5      & 55.3      & 36.3      & 6.0       & 35.8      & 67.4      & 42.7      & 34.6      & 26.1      \\
			Cascade R-CNN~\cite{Cascade_R-CNN_CVPR2018}                        & \TB{36.4} & \TB{56.3} & \TB{38.6} & 6.5       & \TB{38.1} & 70.6      & \TB{44.5} & \TB{36.3} & \TB{28.5} \\
			RetinaNet~\cite{RetinaNet_ICCV2017}                                & 32.5      & 52.2      & 33.7      & 2.6       & 32.8      & 67.9      & 40.0      & 32.5      & 25.0      \\
			ATSS~\cite{ATSS_CVPR2020}                                          & 35.4      & 56.2      & 37.0      & 6.1       & 36.6      & 69.8      & 43.6      & 35.6      & 27.0      \\
			GFL~\cite{GFL_NeurIPS2020}                                         & 35.7      & 56.0      & 37.1      & 6.2       & 36.7      & \TB{70.7} & 44.0      & 36.0      & 27.1      \\
			DETR~\cite{DETR_ECCV2020}                                          & 17.8      & 35.1      & 15.6      & 0.7       & 10.8      & 48.1      & 24.5      & 16.2      & 12.6      \\
			Deformable DETR~\cite{DeformableDETR_ICLR2021}                     & 32.7      & 55.1      & 34.1      & 6.0       & 32.9      & 66.2      & 39.5      & 33.7      & 24.9      \\
			Sparse R-CNN~\cite{Sparse_R-CNN_CVPR2021}                          & 32.8      & 54.3      & 33.9      & \TB{7.2}  & 33.6      & 65.1      & 38.7      & 33.4      & 26.2      \\ \midrule
			ATSS~\cite{ATSS_CVPR2020}$+$Swin-T~\cite{SwinTransformer_ICCV2021} & 37.2      & 58.6      & 38.9      & 7.2       & 39.3      & 71.1      & 45.4      & 36.9      & 29.3      \\
			ATSS~\cite{ATSS_CVPR2020}$+$ConvNeXt-T~\cite{ConvNeXt_CVPR2022}    & \TB{38.3} & \TB{59.9} & \TB{40.1} & \TB{7.4}  & \TB{40.1} & \TB{72.9} & \TB{46.4} & \TB{37.8} & \TB{30.6} \\ \midrule
			ATSS~\cite{ATSS_CVPR2020}$+$SEPC~\cite{SEPC_CVPR2020}              & 35.0      & 55.3      & 36.5      & 5.8       & 35.5      & 70.5      & 43.5      & 35.3      & 26.3      \\
			ATSS~\cite{ATSS_CVPR2020}$+$DyHead~\cite{DyHead_CVPR2021}          & \TB{37.1} & \TB{57.8} & \TB{39.1} & \TB{6.8}  & \TB{39.1} & \TB{72.0} & \TB{45.4} & \TB{37.1} & \TB{28.9} \\ \midrule
			YOLOX-L~\cite{YOLOX_2021} (USB 1.0)                                & 41.0      & 63.4      & 43.2      & 11.5      & 43.9      & 72.8      & 49.0      & 40.9      & 33.1      \\
			YOLOX-L~\cite{YOLOX_2021} (USB 2.0)                                & \TB{41.6} & \TB{64.1} & \TB{43.9} & \TB{12.0} & \TB{44.4} & 73.5      & \TB{49.3} & \TB{41.3} & \TB{34.1} \\
			\OurOrig                                                           & 38.6      & 59.8      & 40.9      & 7.4       & 41.0      & \TB{74.0} & 46.0      & 37.6      & 32.3      \\
			\OurAugust w/o MStrain                                             & 37.9      & 58.8      & 39.4      & 7.8       & 39.2      & 72.9      & 45.9      & 37.5      & 30.2      \\
			\OurAugust                                                         & 39.0      & 60.2      & 40.4      & 8.3       & 41.7      & 73.3      & 47.1      & 38.7      & 31.0      \\ \bottomrule
		\end{tabularx}}
	\end{center}
	\vspace{-3mm}
	\caption{
		Results on WOD \texttt{f0val}.
	}
	\label{table:waymo_f0_train_f0val_832}

	\setlength{\tabcolsep}{0.15em}
	\renewcommand\arraystretch{0.85}
	\begin{center}
		\scalebox{0.7}{\begin{tabularx}{1.02\textwidth}{l*{10}{>{\centering\arraybackslash}X}}
			\toprule
			Method                                                             & AP        & AP$_{50}$ & AP$_{75}$ & \APS      & \APM      & \APL      & body      & face      & frame     & text      \\ \midrule
			Faster R-CNN~\cite{Faster_R-CNN_NIPS2015}                          & 65.8      & \TB{91.1} & 70.6      & \TB{18.4} & 39.9      & 72.1      & 58.3      & 47.5      & 90.1      & 67.1      \\
			Cascade R-CNN~\cite{Cascade_R-CNN_CVPR2018}                        & \TB{67.6} & 90.6      & \TB{72.0} & 17.9      & \TB{41.9} & 74.3      & 60.8      & \TB{48.2} & \TB{92.5} & 69.0      \\
			RetinaNet~\cite{RetinaNet_ICCV2017}                                & 65.3      & 90.5      & 69.5      & 15.7      & 38.9      & 71.9      & 58.3      & 46.3      & 88.8      & 67.7      \\
			ATSS~\cite{ATSS_CVPR2020}                                          & 66.5      & 90.1      & 70.8      & 16.8      & 38.9      & 74.0      & 60.9      & 44.6      & 91.3      & 69.0      \\
			GFL~\cite{GFL_NeurIPS2020}                                         & 67.3      & 90.6      & 71.5      & 17.9      & 38.9      & \TB{74.4} & \TB{61.7} & 45.7      & 92.2      & \TB{69.4} \\
			DETR~\cite{DETR_ECCV2020}                                          & 31.2      & 63.0      & 27.1      & 1.1       & 8.2       & 41.3      & 24.8      & 16.0      & 65.0      & 19.1      \\
			Deformable DETR~\cite{DeformableDETR_ICLR2021}                     & 64.1      & 90.1      & 67.9      & 16.1      & 34.8      & 71.0      & 57.0      & 45.6      & 89.0      & 64.8      \\
			Sparse R-CNN~\cite{Sparse_R-CNN_CVPR2021}                          & 63.1      & 85.8      & 66.3      & 15.3      & 33.8      & 70.3      & 54.2      & 45.5      & 88.9      & 63.5      \\ \midrule
			ATSS~\cite{ATSS_CVPR2020}$+$Swin-T~\cite{SwinTransformer_ICCV2021} & 66.2      & 90.1      & 70.1      & 16.1      & 39.1      & 73.8      & 60.5      & 44.0      & 91.7      & 68.6      \\
			ATSS~\cite{ATSS_CVPR2020}$+$ConvNeXt-T~\cite{ConvNeXt_CVPR2022}    & \TB{67.4} & \TB{90.8} & \TB{71.5} & 16.5      & \TB{39.8} & \TB{75.0} & \TB{62.9} & \TB{45.3} & \TB{92.1} & \TB{69.2} \\ \midrule
			ATSS~\cite{ATSS_CVPR2020}$+$SEPC~\cite{SEPC_CVPR2020}              & 67.1      & 90.2      & 71.5      & 16.2      & 39.8      & 74.9      & 62.3      & 44.6      & 92.1      & 69.4      \\
			ATSS~\cite{ATSS_CVPR2020}$+$DyHead~\cite{DyHead_CVPR2021}          & \TB{67.9} & \TB{90.6} & \TB{72.3} & \TB{17.1} & \TB{42.9} & \TB{75.5} & \TB{63.4} & \TB{45.3} & \TB{92.7} & \TB{70.1} \\ \midrule
			YOLOX-L~\cite{YOLOX_2021} (USB 1.0)                                & 70.1      & \TB{93.7} & \TB{75.0} & 22.3      & \TB{48.4} & 76.1      & 65.8      & 50.9      & 93.0      & 70.8      \\
			YOLOX-L~\cite{YOLOX_2021} (USB 2.0)                                & \TB{70.2} & 93.6      & \TB{75.0} & \TB{22.6} & 47.5      & 76.1      & 65.9      & \TB{51.0} & 93.1      & 70.7      \\
			\OurOrig                                                           & 68.9      & 91.4      & 73.7      & 18.7      & 43.4      & 76.6      & 65.8      & 46.6      & 93.0      & 70.3      \\
			\OurAugust w/o MStrain                                             & 68.3      & 91.2      & 72.2      & 17.9      & 40.9      & 75.6      & 63.3      & 46.3      & 93.1      & 70.3      \\
			\OurAugust                                                         & 69.9      & 92.5      & 74.3      & 20.5      & 43.6      & \TB{77.1} & \TB{66.6} & 48.0      & \TB{93.7} & \TB{71.2} \\ \bottomrule
		\end{tabularx}}
	\end{center}
	\vspace{-3mm}
	\caption{
		Results on \Mangas \texttt{15test}.
	}
	\label{table:Manga109s_15test}
	\end{minipage}
\end{table}

}%

\clearpage

\begin{table*}[t]
\setlength{\tabcolsep}{1.3mm}
\renewcommand\arraystretch{0.83}
\definecolor{mygray}{gray}{0.5}
\newcommand{\DE}[1]{\small{(#1)}}%
\newcommand{\ES}[1]{\textcolor{mygray}{#1}}%
\begin{center}
\scalebox{0.612}{
\begin{tabular}{lll@{\hspace{-1mm}}cclclcccccc}
	\toprule
	Protocol         & Method                                                  & Backbone             & DCN &  Epoch  & Max test scale           &   TTA   & FPS          &    AP     & AP$_{50}$ & AP$_{75}$ &   \APS    &   \APM    &   \APL    \\ \midrule
	Standard USB 1.0 & Faster R-CNN~\cite{Faster_R-CNN_NIPS2015, FPN_CVPR2017} & ResNet-101           &     &   22    & \ES{1333}$\times$\;\:800 &         & \DE{14.2}    &   36.2    &   59.1    &   39.0    &   18.2    &   39.0    &   48.2    \\
	Standard USB 1.0 & Cascade R-CNN~\cite{Cascade_R-CNN_CVPR2018}             & ResNet-101           &     &   19    & 1312$\times$\;\:800      &         & \DE{11.9}    &   42.8    &   62.1    &   46.3    &   23.7    &   45.5    &   55.2    \\
	Standard USB 1.0 & RetinaNet~\cite{RetinaNet_ICCV2017}                     & ResNet-101           &     &   18    & 1333$\times$\;\:800      &         & \DE{13.6}    &   39.1    &   59.1    &   42.3    &   21.8    &   42.7    &   50.2    \\
	Standard USB 1.0 & FCOS~\cite{FCOS_ICCV2019}                               & X-101 (64$\times$4d) &     &   24    & 1333$\times$\;\:800      &         & \DE{\;\:8.9} &   44.7    &   64.1    &   48.4    &   27.6    &   47.5    &   55.6    \\
	Standard USB 1.0 & ATSS~\cite{ATSS_CVPR2020}                               & X-101 (64$\times$4d) & \cm &   24    & 1333$\times$\;\:800      &         & 10.6         &   47.7    &   66.5    &   51.9    &   29.7    &   50.8    &   59.4    \\
	Standard USB 1.0 & FreeAnchor+SEPC~\cite{SEPC_CVPR2020}                    & X-101 (64$\times$4d) & \cm &   24    & 1333$\times$\;\:800      &         & ---          &   50.1    &   69.8    &   54.3    &   31.3    &   53.3    &   63.7    \\
	Standard USB 1.0 & PAA~\cite{PAA_ECCV2020}                                 & X-101 (64$\times$4d) & \cm &   24    & 1333$\times$\;\:800      &         & ---          &   49.0    &   67.8    &   53.3    &   30.2    &   52.8    &   62.2    \\
	Standard USB 1.0 & PAA~\cite{PAA_ECCV2020}                                 & X-152 (32$\times$8d) & \cm &   24    & 1333$\times$\;\:800      &         & ---          &   50.8    &   69.7    &   55.1    &   31.4    &   54.7    &   65.2    \\
	Standard USB 1.0 & RepPoints v2~\cite{RepPointsv2_NeurIPS2020}             & X-101 (64$\times$4d) & \cm &   24    & 1333$\times$\;\:800      &         & \DE{\;\:3.8} &   49.4    &   68.9    &   53.4    &   30.3    &   52.1    &   62.3    \\
	Standard USB 1.0 & RelationNet++~\cite{RelationNet2_NeurIPS2020}           & X-101 (64$\times$4d) & \cm &   20    & 1333$\times$\;\:800      &         & 10.3         &   50.3    &   69.0    &   55.0    & \TB{32.8} &   55.0    & \TB{65.8} \\
	Standard USB 1.0 & GFL~\cite{GFL_NeurIPS2020}                              & ResNet-50            &     &   24    & 1333$\times$\;\:800      &         & 37.2         &   43.1    &   62.0    &   46.8    &   26.0    &   46.7    &   52.3    \\
	Standard USB 1.0 & GFL~\cite{GFL_NeurIPS2020}                              & ResNet-101           &     &   24    & 1333$\times$\;\:800      &         & 29.5         &   45.0    &   63.7    &   48.9    &   27.2    &   48.8    &   54.5    \\
	Standard USB 1.0 & GFL~\cite{GFL_NeurIPS2020}                              & ResNet-101           & \cm &   24    & 1333$\times$\;\:800      &         & 22.8         &   47.3    &   66.3    &   51.4    &   28.0    &   51.1    &   59.2    \\
	Standard USB 1.0 & GFL~\cite{GFL_NeurIPS2020}                              & X-101 (32$\times$4d) & \cm &   24    & 1333$\times$\;\:800      &         & 15.4         &   48.2    &   67.4    &   52.6    &   29.2    &   51.7    &   60.2    \\
	Standard USB 1.0 & \OurAugustS                                             & ResNet-50-C          & \cm &   24    & 1333$\times$\;\:800      &         & 31.6         &   47.4    &   66.0    &   51.4    &   28.3    &   50.8    &   59.5    \\
	Standard USB 1.0 & \OurAugust                                              & Res2Net-50-v1b       & \cm &   24    & 1333$\times$\;\:800      &         & 24.9         &   48.8    &   67.5    &   53.0    &   30.1    &   52.3    &   61.1    \\
	Standard USB 1.0 & \OurAugustD                                             & Res2Net-101-v1b      & \cm &   20    & 1333$\times$\;\:800      &         & 11.7         & \TB{51.3} & \TB{70.0} & \TB{55.8} &   31.7    & \TB{55.3} &   64.9    \\ \midrule
	Large USB 1.0    & \OurAugustD                                             & Res2Net-101-v1b      & \cm &   20    & 1493$\times$\;\:896      &         & 11.6         &   51.5    &   70.2    &   56.0    &   32.8    &   55.5    &   63.7    \\
	Large USB 1.0    & \OurAugustD                                             & Res2Net-101-v1b      & \cm &   20    & 2000$\times$1200         &    5    & ---          & \TB{53.8} & \TB{71.5} & \TB{59.4} & \TB{35.3} & \TB{57.3} & \TB{67.3} \\ \midrule
	Huge USB 1.0     & ATSS~\cite{ATSS_CVPR2020}                               & X-101 (64$\times$4d) & \cm &   24    & 3000$\times$1800         &   13    & ---          &   50.7    &   68.9    &   56.3    &   33.2    &   52.9    &   62.4    \\
	Huge USB 1.0     & PAA~\cite{PAA_ECCV2020}                                 & X-101 (64$\times$4d) & \cm &   24    & 3000$\times$1800         &   13    & ---          &   51.4    &   69.7    &   57.0    &   34.0    &   53.8    &   64.0    \\
	Huge USB 1.0     & PAA~\cite{PAA_ECCV2020}                                 & X-152 (32$\times$8d) & \cm &   24    & 3000$\times$1800         &   13    & ---          &   53.5    & \TB{71.6} &   59.1    & \TB{36.0} &   56.3    &   66.9    \\
	Huge USB 1.0     & RepPoints v2~\cite{RepPointsv2_NeurIPS2020}             & X-101 (64$\times$4d) & \cm &   24    & \ES{3000$\times$1800}    & \ES{13} & ---          &   52.1    &   70.1    &   57.5    &   34.5    &   54.6    &   63.6    \\
	Huge USB 1.0     & RelationNet++~\cite{RelationNet2_NeurIPS2020}           & X-101 (64$\times$4d) & \cm &   20    & \ES{3000$\times$1800}    & \ES{13} & ---          &   52.7    &   70.4    &   58.3    &   35.8    &   55.3    &   64.7    \\
	Huge USB 1.0     & \OurAugustD                                             & Res2Net-101-v1b      & \cm &   20    & 3000$\times$1800         &   13    & ---          & \TB{54.1} & \TB{71.6} & \TB{59.9} &   35.8    &   57.2    & \TB{67.4} \\ \midrule
	Huge USB 2.0     & TSD~\cite{TSD_CVPR2020}                                 & SENet-154            & \cm & \ES{34} & \ES{2000$\times$1400}    & \ES{4}  & ---          &   51.2    & \TB{71.9} &   56.0    &   33.8    &   54.8    &   64.2    \\ \midrule
	Huge USB 2.5     & DetectoRS~\cite{DetectoRS_2020}                         & X-101 (32$\times$4d) & \cm &   40    & 2400$\times$1600         &    5    & ---          & \TB{54.7} & \TB{73.5} & \TB{60.1} & \TB{37.4} & \TB{57.3} &   66.4    \\ \midrule
	Mini USB 3.0     & EfficientDet-D0~\cite{EfficientDet_CVPR2020}            & EfficientNet-B0      &     &   300   & \;\:512$\times$\;\:512   &         & 98.0         &   33.8    &   52.2    &   35.8    &   12.0    &   38.3    &   51.2    \\
	Mini USB 3.1     & YOLOv4~\cite{YOLOv4_2020}                               & CSPDarknet-53        &     &   273   & \;\:512$\times$\;\:512   &         & 83           & \TB{43.0} & \TB{64.9} & \TB{46.5} & \TB{24.3} & \TB{46.1} & \TB{55.2} \\ \midrule
	Standard USB 3.0 & EfficientDet-D2~\cite{EfficientDet_CVPR2020}            & EfficientNet-B2      &     &   300   & \;\:768$\times$\;\:768   &         & 56.5         &   43.0    &   62.3    &   46.2    &   22.5    &   47.0    &   58.4    \\
	Standard USB 3.0 & EfficientDet-D4~\cite{EfficientDet_CVPR2020}            & EfficientNet-B4      &     &   300   & 1024$\times$1024         &         & 23.4         &   49.4    &   69.0    &   53.4    &   30.3    &   53.2    &   63.2    \\
	Standard USB 3.1 & YOLOv4~\cite{YOLOv4_2020}                               & CSPDarknet-53        &     &   273   & \;\:608$\times$\;\:608   &         & 62           &   43.5    &   65.7    &   47.3    &   26.7    &   46.7    &   53.3    \\ \midrule
	Large USB 3.0    & EfficientDet-D5~\cite{EfficientDet_CVPR2020}            & EfficientNet-B5      &     &   300   & 1280$\times$1280         &         & 13.8         &   50.7    &   70.2    &   54.7    &   33.2    &   53.9    &   63.2    \\
	Large USB 3.0    & EfficientDet-D6~\cite{EfficientDet_CVPR2020}            & EfficientNet-B6      &     &   300   & 1280$\times$1280         &         & 10.8         &   51.7    &   71.2    &   56.0    &   34.1    &   55.2    &   64.1    \\
	Large USB 3.0    & EfficientDet-D7~\cite{EfficientDet_CVPR2020}            & EfficientNet-B6      &     &   300   & 1536$\times$1536         &         & \;\:8.2      &   52.2    &   71.4    &   56.3    &   34.8    &   55.5    &   64.6    \\ \midrule
	Freestyle        & RetinaNet+SpineNet~\cite{SpineNet_CVPR2020}             & SpineNet-190         &     &   400   & 1280$\times$1280         &         & ---          &   52.1    &   71.8    &   56.5    &   35.4    &   55.0    &   63.6    \\
	Freestyle        & EfficientDet-D7x~\cite{EfficientDet_arXiv}              & EfficientNet-B7      &     &   600   & 1536$\times$1536         &         & \;\:6.5      & \TB{55.1} & \TB{74.3} &   59.9    &   37.2    & \TB{57.9} & \TB{68.0} \\ \bottomrule
\end{tabular}
}
\end{center}
\vspace{-2.5mm}
\caption{
	State-of-the-art methods on COCO \texttt{test-dev}.
	We classify methods by the proposed protocols without compatibility.
	X in the Backbone column denotes ResNeXt~\cite{ResNeXt_CVPR2017}. See method papers for other backbones.
	TTA: Test-time augmentation including horizontal flip and multi-scale testing (numbers denote scales).
	FPS values without and with parentheses were measured on V100 with mixed precision and other environments, respectively.
	We measured the FPS of GFL~\cite{GFL_NeurIPS2020} models in our environment
	and estimated those of ATSS~\cite{ATSS_CVPR2020} and RelationNet++~\cite{RelationNet2_NeurIPS2020} based on the measured values and \cite{GFL_NeurIPS2020, RelationNet2_NeurIPS2020}.
	Other methods' settings are based on conference papers, their arXiv versions, and authors' codes.
	Values shown in gray were estimated from descriptions in papers and codes.
	Some FPS values are from~\cite{GFL_NeurIPS2020}.
}
\label{table:coco_sota}
\end{table*}

\subsection{Rethinking COCO with USB Protocols}
\label{sec:rethink_coco_sota}

We classify state-of-the-art methods on COCO \texttt{test-dev} (as of November 14, 2020)
by the proposed protocols without compatibility.
The results are shown in Table~\ref{table:coco_sota}.
Although state-of-the-art detectors on the COCO benchmark were trained with various settings,
the introduced divisions enable us to compare methods in each division.
\OurAugustD achieves the highest AP (51.3\%) in the Standard USB 1.0 protocol.
Despite 12.5$\times$ fewer epochs, the speed-accuracy trade-offs of \Univs are comparable to those of EfficientDet~\cite{EfficientDet_CVPR2020}.
With 13-scale TTA, \OurAugustD achieves the highest AP (54.1\%) in the Huge USB 1.0 protocol.
Results in higher protocols than USB 1.0 are scattered.
If we ignore the difference of hyperparameter optimization,
YOLOv4~\cite{YOLOv4_2020} shows a better speed-accuracy trade-off than EfficientDet~\cite{EfficientDet_CVPR2020} in Standard USB 3.x.

Comparisons across different divisions are difficult.
Especially, long training is problematic because it can secretly increase AP without decreasing FPS, unlike large test scales.
Nevertheless, the EfficientDet~\cite{EfficientDet_CVPR2020}, YOLOv4~\cite{YOLOv4_2020}, and SpineNet~\cite{SpineNet_CVPR2020} papers
compare methods in their tables without specifying the difference in training epochs.
The compatibility of the USB training protocols resolves this disorder.
We hope that many papers report results with the protocols for inclusive, healthy, and sustainable development of detectors.

To simulate the compatibility from Standard USB 3.0 to 1.0,
we refer to the training log of the EfficientDet author.
The AP of EfficientDet-D4~\cite{EfficientDet_CVPR2020} on COCO \texttt{minival} is 43.8\% at 23 epoch~\cite{EfficientDetGitHubComment}.
Although it could be improved by changing the learning rate schedule,
EfficientDet's inference efficiency is not compatible with training efficiency.

\begin{table}[t]
	\setlength{\tabcolsep}{2.3mm}
	\renewcommand\arraystretch{0.85}
	\newcommand{\RANK}{\multirow{2}{*}[-0.5\dimexpr \aboverulesep + \belowrulesep + \cmidrulewidth]{Rank}}
	\newcommand{\METHOD}{\multirow{2}{*}[-0.5\dimexpr \aboverulesep + \belowrulesep + \cmidrulewidth]{Method}}
	\newcommand{\APLL}{\multirow{2}{*}[-0.5\dimexpr \aboverulesep + \belowrulesep + \cmidrulewidth]{AP/L2}}
	\newcommand{\CMRs}{\cmidrule(l{.2em}r{.2em}){3-4}}
	\begin{center}
		\scalebox{0.7}{\begin{tabular}{llccc}
				\toprule
				\RANK & \METHOD                                                             & \multicolumn{2}{c}{\# Models} &     \APLL      \\
				\CMRs &                                                                     & Multi-stage & Single-stage    &                \\ \midrule
				\multicolumn{5}{l}{\textit{Methods including multi-stage detector:}}                                                         \\
				1     & RW-TSDet~\cite{Waymo2d_1st_2020}                                    & 6+          &                 & \textbf{74.43} \\
				2     & HorizonDet~\cite{Waymo2d_2nd_2020}                                  & 4           & 8               &     70.28      \\
				3     & SPNAS-Noah~\cite{Waymo2d_3rd_2020}                                  & 2           &                 &     69.43      \\ \midrule
				\multicolumn{5}{l}{\textit{Single-stage detectors:}}                                                                         \\
				7     & \textbf{UniverseNet (Ours)}                                         &             & \textbf{1}      & \textbf{67.42} \\
				13    & YOLO V4~\cite{YOLOv4_2020}                                          &             & 1+              &     58.08      \\
				14    & ATSS-Efficientnet~\cite{ATSS_CVPR2020, EfficientNet_ICML2019}       &             & 1+              &     56.99      \\ \bottomrule
		\end{tabular}}
	\end{center}
	\vspace{-3mm}
	\caption{
		Waymo Open Dataset Challenge 2020 2D detection~\cite{WaymoOpenDataset_2D_detection_leaderboard}.
	}
	\label{table:waymo_leaderboard}
\end{table}

\subsection{Comparison with State-of-the-Art}
\label{sec:sota_comparison}

\noindent
\textbf{WOD.}
For comparison with state-of-the-art methods on WOD,
we submitted the detection results of \OurOrig to the Waymo Open Dataset Challenge 2020 2D detection,
a competition held at a CVPR 2020 workshop.
The primary metric is AP/L2,
a KITTI-style AP evaluated with LEVEL\_2 objects~\cite{WaymoOpenDataset_CVPR2020, WaymoOpenDataset_2D_detection_leaderboard}.
We used multi-scale testing with soft-NMS~\cite{SoftNMS_ICCV2017}.
The shorter side pixels of test scales are $(960, 1600, 2240)$, including 8 pixels of padding.
These scales enable utilizing SEPC~\cite{SEPC_CVPR2020} (see Sec.~\ref{sec:test_scales}) and detecting small objects.
Table~\ref{table:waymo_leaderboard} shows the top teams' results.
\OurOrig achieves 67.42\% AP/L2
without multi-stage detectors, ensembles, expert models, or heavy backbones, unlike other top methods.
RW-TSDet~\cite{Waymo2d_1st_2020} overwhelms other multi-stage detectors,
whereas UniverseNet overwhelms other single-stage detectors.
These two methods used light backbones and large test scales~\cite{ashraf2016shallow}.
Interestingly, the maximum test scales are the same (3360$\times$2240).
We conjecture that this is not a coincidence but a convergence caused by searching the accuracy saturation point.

\noindent
\textbf{\Mangas.}
To the best of our knowledge, no prior work has reported detection results on the \textit{\Mangas} dataset (87 volumes).
Although many settings differ,
the state-of-the-art method on the full \textit{Manga109} dataset (109 volumes, non-public to commercial organizations)
achieves 77.1--92.0\% (mean: 84.2\%) AP$_{50}$ on ten test volumes~\cite{Manga109_detection_Ogawa_2018}.
The mean AP$_{50}$ of \OurAugust on the \texttt{15test} set (92.5\%) is higher than those results.

\begin{table*}[t]
	\captionsetup[sub]{font={footnotesize,color=captioncolor},width=0.985\linewidth}
	\setlength{\tabcolsep}{1.2mm}
	\renewcommand\arraystretch{0.85}
	\begin{minipage}[c]{0.46\hsize}
		\begin{center}
			\scalebox{0.655}{\begin{tabular}{p{29mm}cccccc}
					\toprule
					Method                                      &    AP     & AP$_{50}$ & AP$_{75}$ &   \APS    &   \APM    &   \APL    \\ \midrule
					ATSS~\cite{ATSS_CVPR2020}                   &   39.4    &   57.6    &   42.8    &   23.6    &   42.9    &   50.3    \\
					\ATSEPC~\cite{ATSS_CVPR2020, SEPC_CVPR2020} & \TB{42.1} & \TB{59.9} & \TB{45.5} & \TB{24.6} & \TB{46.1} & \TB{55.0} \\ \bottomrule
			\end{tabular}}
		\end{center}
		\vspace{-5mm}
		\subcaption{
			AP improvements by SEPC without iBN~\cite{SEPC_CVPR2020}.
		}
		\label{table:coco_ATSEPC}
		\vspace{-2mm}
		\begin{center}
			\scalebox{0.655}{\begin{tabular}{p{29mm}cccccc}
					\toprule
					Method                                      &    AP     & AP$_{50}$ & AP$_{75}$ &   \APS    &   \APM    &   \APL    \\ \midrule
					\ATSEPC~\cite{ATSS_CVPR2020, SEPC_CVPR2020} &   42.1    &   59.9    &   45.5    &   24.6    &   46.1    &   55.0    \\
					\OurOrig                                    & \TB{46.7} & \TB{65.0} & \TB{50.7} & \TB{29.2} & \TB{50.6} & \TB{61.4} \\ \bottomrule
			\end{tabular}}
		\end{center}
		\vspace{-5mm}
		\subcaption{
			AP improvements by Res2Net-v1b~\cite{Res2Net_TPAMI2020}, DCN~\cite{DCN_ICCV2017}, and multi-scale training.
		}
		\label{table:coco_OurOrig}
		\vspace{-2mm}
		\begin{center}
			\scalebox{0.655}{\begin{tabular}{p{29mm}cccccc}
					\toprule
					Method                     &    AP     & AP$_{50}$ & AP$_{75}$ &   \APS    &   \APM    &   \APL    \\ \midrule
					\OurOrig                   &   46.7    &   65.0    &   50.7    & \TB{29.2} &   50.6    &   61.4    \\
					\OurGFL                    & \TB{47.5} & \TB{65.8} & \TB{51.8} & \TB{29.2} & \TB{51.6} & \TB{62.5} \\ \bottomrule
			\end{tabular}}
		\end{center}
		\vspace{-5mm}
		\subcaption{
			AP improvements by GFL~\cite{GFL_NeurIPS2020}.
		}
		\label{table:coco_OurGFL}
		\vspace{-2mm}
		\begin{center}
			\scalebox{0.655}{\begin{tabular}{p{29mm}cccccc}
					\toprule
					Method      &    AP     & AP$_{50}$ & AP$_{75}$ &   \APS    &   \APM    &   \APL    \\ \midrule
					\OurGFL     &   47.5    &   65.8    &   51.8    &   29.2    &   51.6    &   62.5    \\
					\OurAugustD & \TB{48.6} & \TB{67.1} & \TB{52.7} & \TB{30.1} & \TB{53.0} & \TB{63.8} \\ \bottomrule
			\end{tabular}}
		\end{center}
		\vspace{-5mm}
		\subcaption{
			AP improvements by SyncBN~\cite{MegDet_CVPR2018} and iBN~\cite{SEPC_CVPR2020}.
		}
		\label{table:coco_OurAugustD}
	\end{minipage}
	\hfill
	\begin{minipage}[c]{0.54\hsize}
		\begin{center}
			\scalebox{0.655}{\begin{tabular}{lcccccccc}
					\toprule
					Method      &  DCN  &    FPS    &    AP     & AP$_{50}$ & AP$_{75}$ &   \APS    &   \APM    &   \APL    \\ \midrule
					\OurAugustD & heavy &   17.3    & \TB{48.6} & \TB{67.1} & \TB{52.7} & \TB{30.1} & \TB{53.0} & \TB{63.8} \\
					\OurAugust  & light & \TB{24.9} &   47.5    &   66.0    &   51.9    &   28.9    &   52.1    &   61.9    \\ \bottomrule
			\end{tabular}}
		\end{center}
		\vspace{-5mm}
		\subcaption{
			Speeding up by the light use of DCN~\cite{DCN_ICCV2017, SEPC_CVPR2020}.
		}
		\label{table:coco_OurAugust}
		\vspace{-1.5mm}
		\begin{center}
			\scalebox{0.655}{\begin{tabular}{p{38mm}ccccccc}
				\toprule
				Method                                                &    FPS    &    AP     & AP$_{50}$ & AP$_{75}$ &   \APS    &   \APM    &   \APL    \\ \midrule
				\OurAugust                                            &   24.9    & \TB{47.5} & \TB{66.0} & \TB{51.9} & \TB{28.9} & \TB{52.1} & \TB{61.9} \\
				w/o SEPC~\cite{SEPC_CVPR2020}                         &   26.7    &   45.8    &   64.6    &   50.0    &   27.6    &   50.4    &   59.7    \\
				w/o Res2Net-v1b~\cite{Res2Net_TPAMI2020}              & \TB{32.8} &   44.7    &   62.8    &   48.4    &   27.1    &   48.8    &   59.5    \\
				w/o DCN~\cite{DCN_ICCV2017}                           &   27.8    &   45.9    &   64.5    &   49.8    & \TB{28.9} &   49.9    &   59.0    \\
				w/o multi-scale training                              &   24.8    &   45.9    &   64.5    &   49.6    &   27.4    &   50.5    &   60.1    \\
				w/o SyncBN, iBN~\cite{MegDet_CVPR2018, SEPC_CVPR2020} &   25.7    &   45.8    &   64.0    &   50.2    &   27.9    &   50.0    &   59.8    \\ \bottomrule
			\end{tabular}}
		\end{center}
		\vspace{-5mm}
		\subcaption{
			Ablation from \OurAugust.
			Replacing Res2Net-v1b backbone with ResNet-B~\cite{BagOfTricks_Classification_CVPR2019} has the largest effects.
		}
		\label{table:coco_OurAugust_ablation}
		\vspace{-1.5mm}
		\begin{center}
			\scalebox{0.655}{\begin{tabular}{p{38mm}ccccccc}
				\toprule
				Backbone                                               &    FPS    &    AP     & AP$_{50}$ & AP$_{75}$ &   \APS    &   \APM    &   \APL    \\ \midrule
				ResNet-50-B~\cite{BagOfTricks_Classification_CVPR2019} & \TB{32.8} &   44.7    &   62.8    &   48.4    &   27.1    &   48.8    &   59.5    \\
				ResNet-50-C~\cite{BagOfTricks_Classification_CVPR2019} &   32.4    &   45.8    &   64.2    &   50.0    &   28.8    &   50.1    &   60.0    \\
				Res2Net-50~\cite{Res2Net_TPAMI2020}                    &   25.0    &   46.3    &   64.7    &   50.3    &   28.2    &   50.6    &   60.8    \\
				Res2Net-50-v1b~\cite{Res2Net_TPAMI2020}                &   24.9    & \TB{47.5} & \TB{66.0} & \TB{51.9} & \TB{28.9} & \TB{52.1} & \TB{61.9} \\ \bottomrule
			\end{tabular}}
		\end{center}
		\vspace{-5mm}
		\subcaption{
			\OurAugust with different backbones.
		}
		\label{table:coco_backbone}
	\end{minipage}
	\caption{
		Ablation studies on COCO \texttt{minival}.
	}
	\label{table:coco_ablation}
\end{table*}

\subsection{Ablation Studies for \Univs}
\label{sec:coco_ablation}

We show the results of ablation studies for \Univs on COCO in Table~\ref{table:coco_ablation}.
As shown in Table~\ref{table:coco_ATSEPC}, \ATSEPC (ATSS~\cite{ATSS_CVPR2020} with SEPC without iBN~\cite{SEPC_CVPR2020}) outperforms ATSS by a large margin.
The effectiveness of SEPC for ATSS is consistent with those for other detectors reported in the SEPC paper~\cite{SEPC_CVPR2020}.
As shown in Table~\ref{table:coco_OurOrig}, \OurOrig further improves AP metrics by $\sim$5\%
by adopting Res2Net-v1b~\cite{Res2Net_TPAMI2020}, DCN~\cite{DCN_ICCV2017}, and multi-scale training.
As shown in Table~\ref{table:coco_OurGFL}, adopting GFL~\cite{GFL_NeurIPS2020} improves AP by 0.8\%.
There is room for improvement of \APS in the Quality Focal Loss of GFL~\cite{GFL_NeurIPS2020}.
As shown in Table~\ref{table:coco_OurAugustD},
\OurAugustD achieves 48.6\% AP by making more use of BatchNorm (SyncBN~\cite{MegDet_CVPR2018} and iBN~\cite{SEPC_CVPR2020}).
It is much more accurate than other models trained for 12 epochs using ResNet-50-level backbones (\eg, ATSS: 39.4\%~\cite{ATSS_CVPR2020, MMDetection}, GFL: 40.2\%~\cite{GFL_NeurIPS2020, MMDetection}).
On the other hand, the inference is not so fast (less than 20 FPS) due to the heavy use of DCN~\cite{DCN_ICCV2017}.
\OurAugust speeds up inference by the light use of DCN~\cite{DCN_ICCV2017, SEPC_CVPR2020}.
As shown in Table~\ref{table:coco_OurAugust},
\OurAugust is 1.4$\times$ faster than \OurAugustD at the cost of a $\sim$1\% AP drop.
To further verify the effectiveness of each technique, we conducted ablation from \OurAugust shown in Table~\ref{table:coco_OurAugust_ablation}.
All techniques contribute to the high AP of \OurAugust.
Ablating the Res2Net-v1b backbone (replacing Res2Net-50-v1b~\cite{Res2Net_TPAMI2020} with ResNet-50-B~\cite{BagOfTricks_Classification_CVPR2019}) has the largest effects.
Res2Net-v1b improves AP by 2.8\% and increases the inference time by 1.3$\times$.
To further investigate the effectiveness of backbones, we trained variants of \OurAugust as shown in Table~\ref{table:coco_backbone}.
Although the Res2Net module~\cite{Res2Net_TPAMI2020} makes inference slower,
the deep stem used in ResNet-50-C~\cite{BagOfTricks_Classification_CVPR2019} and Res2Net-50-v1b~\cite{Res2Net_TPAMI2020}
improves AP metrics with similar speeds.
\OurAugustS (the variant using the ResNet-50-C backbone) shows a good speed-accuracy trade-off by achieving 45.8\% AP and over 30 FPS.

\subsection{Effects of Test Scales}
\label{sec:test_scales}

\begin{figure}[t]
	\centering
	\begin{minipage}[b]{0.48\linewidth}
		\centering
		\includegraphics[width=\linewidth]{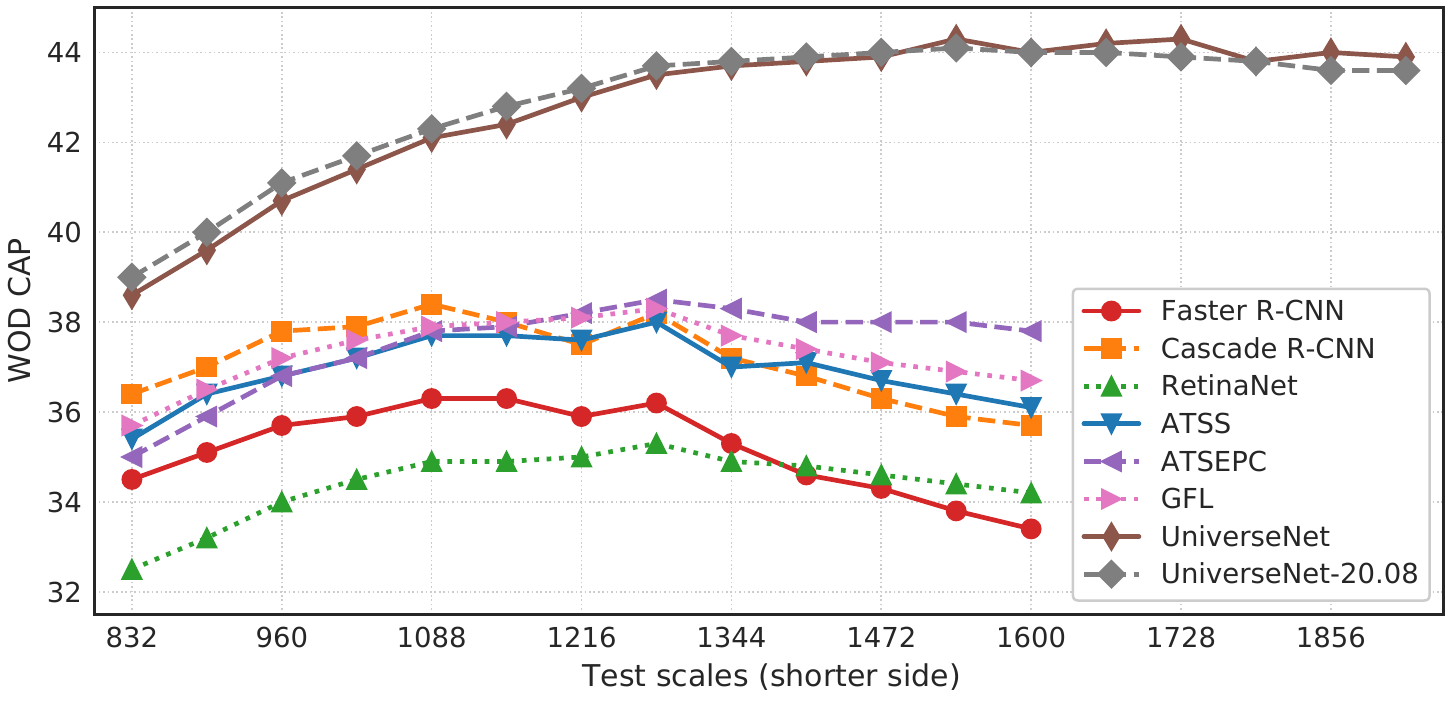}
		\vspace{-6mm}
		\subcaption{COCO-style AP}
		\label{fig:waymo_scales_cap_supmat}
	\end{minipage}
	\hfill
	\begin{minipage}[b]{0.48\linewidth}
		\centering
		\includegraphics[width=\linewidth]{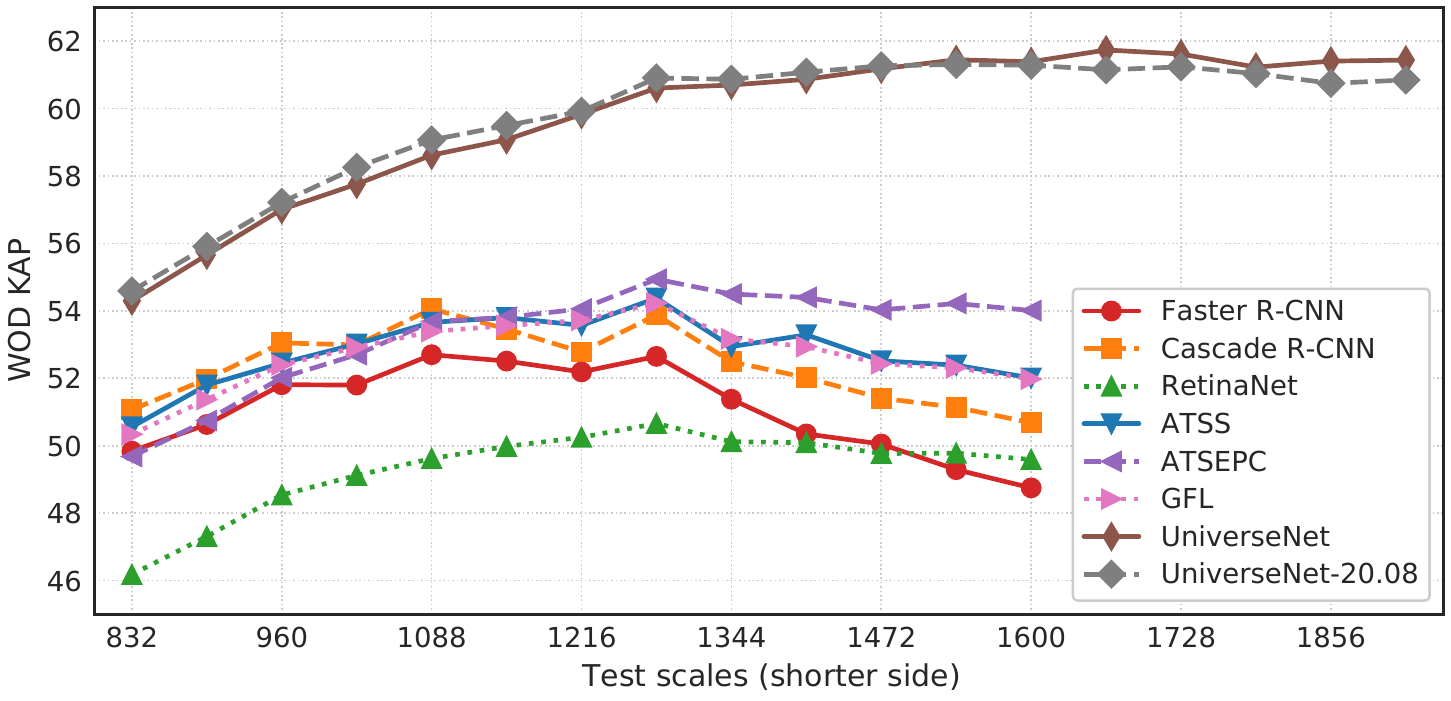}
		\vspace{-6mm}
		\subcaption{KITTI-style AP}
		\label{fig:waymo_scales_kap_supmat}
	\end{minipage}
	\caption{
		Test scales \vs different AP metrics on WOD \texttt{f0val}.
	}
	\label{fig:waymo_scales_supmat}
\end{figure}

We show the results on WOD at different test scales in Figure~\ref{fig:waymo_scales_cap_supmat}.
Single-stage detectors require larger test scales than multi-stage detectors to achieve peak performance,
probably because they cannot extract features from precisely localized region proposals.
Although \ATSEPC shows lower AP than ATSS at the default test scale (1248$\times$832 in Standard USB),
it outperforms ATSS at larger test scales (\eg, 1920$\times$1280 in Large USB).
We conjecture that we should enlarge object scales in images to utilize SEPC~\cite{SEPC_CVPR2020}
because its DCN~\cite{DCN_ICCV2017} enlarges effective receptive fields.
SEPC and DCN prefer large objects empirically (Tables~\ref{table:coco_ATSEPC}, \ref{table:coco_OurAugust_ablation}, \cite{SEPC_CVPR2020, DCN_ICCV2017}),
and DCN~\cite{DCN_ICCV2017} cannot increase the sampling points for objects smaller than the kernel size in principle.
By utilizing the characteristics of SEPC and multi-scale training,
\Univs achieve the highest AP in a wide range of test scales.

\subsection{Evaluation with KITTI-Style AP}

We evaluated the KITTI-style AP (KAP) on WOD.
KAP is a metric used in benchmarks for autonomous driving~\cite{KITTI_CVPR2012, WaymoOpenDataset_CVPR2020}.
Using different IoU thresholds (0.7 for vehicles, and 0.5 for pedestrians and cyclists),
KAP is calculated as
$\mathrm{KAP} = (\mathrm{AP_{0.7, veh.}}+\mathrm{AP_{0.5, ped.}}+\mathrm{AP_{0.5, cyc.}}) / 3.$
The results of KAP are shown in Figure~\ref{fig:waymo_scales_kap_supmat}.
GFL~\cite{GFL_NeurIPS2020} and Cascade R-CNN~\cite{Cascade_R-CNN_CVPR2018}, which focus on localization quality, are less effective for KAP.

\subsection{Effects of COCO Pre-Training}

To verify the effects of COCO pre-training,
we trained \OurAugust on \MangasAbbr from different pre-trained models.
Table~\ref{table:Manga109s_15test_pretrain} shows the results.
COCO pre-training improves all the metrics, especially body AP.

\begin{table}[t]
	\setlength{\tabcolsep}{0.9mm}
	\renewcommand\arraystretch{0.72}
	\begin{center}
		\scalebox{0.7}{\begin{tabular}{lcccccccccc}
			\toprule
			Pre-training   &    AP     & AP$_{50}$ & AP$_{75}$ &   \APS    &   \APM    &   \APL    &   body    &   face    &   frame   &   text    \\ \midrule
			ImageNet       &   68.9    &   92.2    &   73.3    &   19.9    &   42.6    &   75.8    &   64.3    &   47.6    &   93.0    &   70.7    \\
			COCO 1$\times$ & \TB{69.9} & \TB{92.5} & \TB{74.3} & \TB{20.5} & \TB{43.6} & \TB{77.1} & \TB{66.6} & \TB{48.0} &   93.7    & \TB{71.2} \\
			COCO 2$\times$ &   69.8    &   92.3    &   74.0    & \TB{20.5} &   43.4    &   77.0    &   66.5    &   47.8    & \TB{93.8} & \TB{71.2} \\ \bottomrule
		\end{tabular}}
	\end{center}
	\vspace{-3mm}
	\caption{
		\OurAugust on Manga109-s \texttt{15test} with different pre-trained models.
	}
	\label{table:Manga109s_15test_pretrain}
\end{table}

\end{document}